\pdfoutput=1
\documentclass[11pt]{article}

\usepackage[table]{xcolor}
\usepackage[final]{acl}

\usepackage{times}
\usepackage{latexsym}
\usepackage{multirow}

\usepackage[T1]{fontenc}
\usepackage[utf8]{inputenc}
\usepackage{booktabs}
\usepackage{microtype}
\usepackage{paralist}

\usepackage{inconsolata}

\usepackage{graphicx}
\usepackage[ruled,vlined]{algorithm2e}
\usepackage{amsmath} 

\title{LLMs as Cultural Archives: \\Cultural Commonsense Knowledge Graph Extraction}
\author{
    Junior Cedric Tonga\textsuperscript{1} \quad
    Chen Cecilia Liu\textsuperscript{2} \quad
    Iryna Gurevych\textsuperscript{1,2} \quad Fajri Koto\textsuperscript{1} \\
    \textsuperscript{1}Mohamed bin Zayed University of Artificial Intelligence\\
    \textsuperscript{2}Ubiquitous Knowledge Processing Lab (UKP Lab)\\
    Department of Computer Science and Hessian Center for AI (hessian.AI)\\
    Technische Universität Darmstadt\\
    \texttt{\small \{junior.tonga, fajri.koto\}@mbzuai.ac.ae} 
}

\begin{document}
\maketitle
\begin{abstract}
Large language models (LLMs) encode rich cultural knowledge learned from diverse web-scale data, offering an unprecedented opportunity to model cultural commonsense at scale. Yet this knowledge remains mostly implicit and unstructured, limiting its interpretability and use. We present an iterative, prompt-based framework for constructing a Cultural Commonsense Knowledge Graph (CCKG) that treats LLMs as cultural archives, systematically eliciting culture-specific entities, relations, and practices and composing them into multi-step inferential chains across languages. We evaluate CCKG on five countries with human judgments of cultural relevance, correctness, and path coherence. We find that the cultural knowledge graphs are better realized in English, even when the target culture is non-English (e.g., Chinese, Indonesian, Arabic), indicating uneven cultural encoding in current LLMs. Augmenting smaller LLMs with CCKG improves performance on cultural reasoning and story generation, with the largest gains from English chains. Our results show both the promise and limits of LLMs as cultural technologies and that chain-structured cultural knowledge is a practical substrate for culturally grounded NLP.\footnote{Code available at \url{https://github.com/JuniorTonga/Cultural_Commonsense_Knowledge_Graph}}

\end{abstract}

\begin{figure}[t!]
    \centering
    \includegraphics[width=\linewidth]{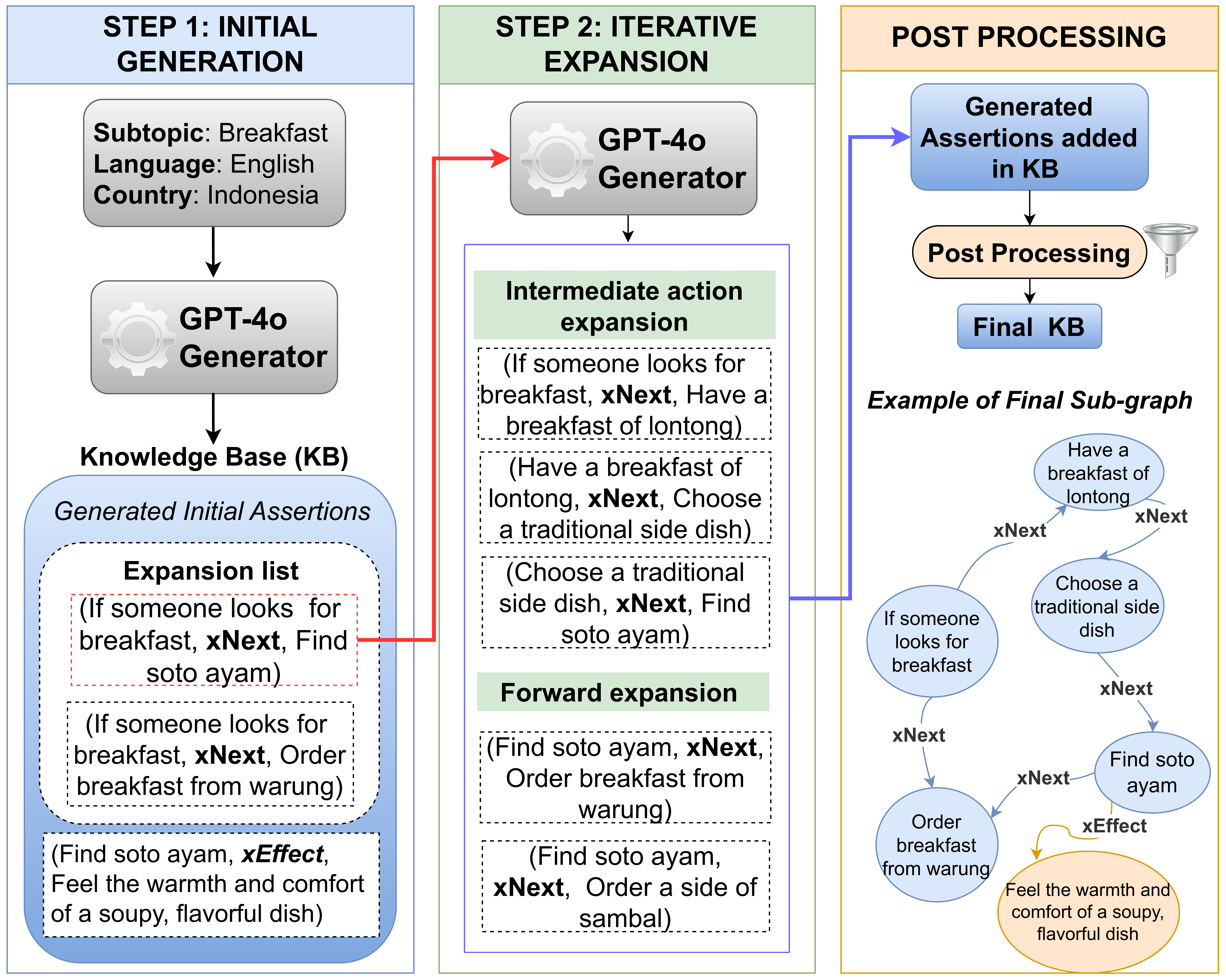}
    \caption{Application of our framework for constructing a partial Cultural Commonsense Knowledge Graph (CCKG) capturing culturally grounded reasoning about breakfast in Indonesia. Given an input prompt specifying the subtopic, language, country, and task-specific constraints, GPT-4o generates English \textit{if--then} commonsense assertions $(\textit{action}_i, \textit{relation}, \textit{action}_j)$ to form an initial knowledge base (KB). Assertions with relations (\texttt{xNext}, \texttt{oNext}) are iteratively expanded by re-prompting GPT-4o to generate \textbf{intermediate action expansions} that decompose $\textit{action}_i$ into finer-grained steps leading to $\textit{action}_j$ and \textbf{forward actions} occurring after $\textit{action}_j$. In this example, only the first assertion in the expansion list is expanded for a single iteration. The resulting assertions are added to the KB, post-processed and composed into the final CCKG subgraph.
}
 \label{fig:main}
    %\vspace{-0.4cm}
\end{figure}

\section{Introduction}

Culture and commonsense reasoning are deeply intertwined, as culture shapes how people interpret everyday situations, social conventions, and causal regularities \cite{tacl/koto-etal-2024-indoculture, acl/sadallah-etal-2025-commonsense, sap-etal-2020-commonsense}. Culture encompasses shared and learned values, norms, and practices that guide interpretation and action within a community \cite{acl/HershcovichFLLA22, emnlp/adilazuarda-etal-2024-towards, tacl/liu-etal-2025-culturally}. Because commonsense is grounded in cultural experience rather than universal logic, what seems self-evident in one community can be unfamiliar, or even misleading, in another \cite{naous-etal-2024-beer, almheiri-etal-2025-cross}. While early computational approaches largely treated commonsense as culture-neutral \cite{DBLP:conf/aaai/BiskZLGC20,sap2019socialiqa}, recent works have begun to recognize and model its cultural dimensions, highlighting the linguistic and cognitive variation that arises across communities \cite{tacl/koto-etal-2024-indoculture, acl/sadallah-etal-2025-commonsense}.

We argue that language models can reason more appropriately across contexts when equipped with structured representations of culture, especially given its complexity and uneven representation in online data. Early efforts such as \texttt{ATOMIC} \cite{aaai/atomic/SapBABLRRSC19} demonstrated the value of modeling inferential knowledge between everyday events \cite{hwang2021cometatomic2020symbolicneural,bosselut-etal-2019-comet}, yet these resources remain largely Western-centric and lack cross-cultural generalizability. Extending this idea to cultural contexts requires uncovering and organizing the implicit cultural knowledge within LLMs into interpretable, multi-step structures that reflect local norms and practices. 

In this work, we study how far large language models serve as cultural archives---examining the extent and accuracy of the cultural information they encode. Since large language models are pretrained on culturally diverse corpora \cite{jiang-etal-2021-know,sun-etal-2024-head,brown2020language}, much of this knowledge already exists implicitly and can be systematically extracted in structured form. Specifically, we address three research questions: (1) \textit{To what extent do LLMs encode cultural knowledge that aligns with real-world cultural relevance?} (2) \textit{Which language best represents cultural knowledge when extracted from LLMs?} and (3) \textit{Can the extracted cultural knowledge be used to enhance the cultural reasoning ability of smaller or weaker models?} While our initial hypothesis assumes that the language spoken by the culture 
provides the most authentic representation, our findings reveal an interesting contrast: English often serves as a more coherent medium for representing cultural knowledge graphs.

Prior work has constructed cultural knowledge bases from sources such as Wikipedia, CommonCrawl, and social media \cite{www/NguyenRVW23-Mango, fung-2024-cultureatlas, emnlp/shi-etal-2024-culturebank}. While valuable, these approaches face two key limitations. First, they represent culture as isolated, static facts \cite{yin2022geomlama, www/NguyenRVW23-Mango, fung-2024-cultureatlas}, overlooking the procedural and contextual nature of cultural practices. Many traditions consist of ordered sequences of actions---such as the proposal-to-marriage process in Indonesian weddings---and reducing them to atomic statements omits crucial context, limiting applications like reasoning and story generation. Second, most cultural knowledge bases are built in English, despite the fact that many cultural nuances are best expressed in native languages.

Our contributions can be summarized as follows:
\begin{compactitem}
    \item We propose an iterative, prompt-based framework for constructing \textbf{Cultural Commonsense Knowledge Graphs (CCKG)}, which extract multilingual, culture-specific \textit{if--then} inferential knowledge chains from large language models. Following the \texttt{ATOMIC}-style formulation \cite{aaai/atomic/SapBABLRRSC19}, we compose these relations into multi-step cultural knowledge chains (Figure~\ref{fig:main}).
    \item Through extensive human evaluations of cultural relevance, correctness, and path coherence, we find that while native languages capture richer cultural detail, English extractions are more coherent and consistently preferred.
    \item We show that augmenting smaller or weaker LLMs with CCKG improves their cultural reasoning and story generation performance, highlighting the value of inferential cultural knowledge for developing culturally grounded NLP systems.
\end{compactitem}

\section{Related work}

\paragraph{Knowledge Bases for Commonsense Reasoning.}
Early commonsense knowledge bases (KBs) such as \texttt{WebChild}~\cite{Tandon2014WebChildHA} linked nouns with descriptive adjectives to encode physical and perceptual properties, providing one of the first large-scale automatically constructed commonsense resources. \texttt{ConceptNet}~\cite{DBLP:conf/aaai/SpeerCH17} aggregated human-authored assertions but primarily captured lexical relations between words rather than inferential or situational knowledge. \texttt{Quasimodo}~\cite{10.1145/3357384.3357955} further expanded coverage by mining commonsense assertions from QA forums and web text through large-scale automated extraction. \texttt{ATOMIC}~\cite{aaai/atomic/SapBABLRRSC19} marked a shift toward event-level reasoning, introducing large-scale \emph{if–then} relations that capture causes, effects, and social intentions.

While these knowledge bases advanced commonsense modeling, they remain largely English-centric and culturally neutral. Recent work has begun integrating culture into knowledge representation: \texttt{CANDLE}~\cite{candle/www/NguyenRVW23} performs large-scale web extraction to probe factual knowledge across regional contexts, \texttt{Mango}~\cite{www/NguyenRVW23-Mango} extracts cross-cultural facts from LLMs, \texttt{CultureAtlas}~\cite{fung-2024-cultureatlas} constructs a cultural knowledge base from Wikipedia and Wikidata, and \texttt{CultureBank}~\cite{emnlp/shi-etal-2024-culturebank} collects cultural descriptors from social media such as TikTok and Reddit. However, these efforts primarily focus on factual or descriptive knowledge—typically short, unstructured snippets without inferential chains or sequential reasoning.

In contrast, we treat large language models as cultural archives, leveraging their internalized knowledge to construct structured, inferential representations. We introduce the Cultural Commonsense Knowledge Graph (CCKG)--a graph-based resource that models \textit{if–then} reasoning chains involving human actions, intentions, and consequences.

\paragraph{Evaluating Cultural Commonsense Knowledge.} 

Early research on commonsense evaluation was predominantly conducted in English, often lacking strong cultural grounding and reflecting universal or Western-centric perspectives. This research spans physical commonsense~\citep{DBLP:conf/aaai/BiskZLGC20}, which evaluates models' understanding of real-world dynamics and object relations, and social reasoning~\citep{sap2019socialiqa}, which tests their ability to infer human emotions, intentions, and social norms. Later studies extended this line of inquiry to numerical~\citep{lin-etal-2020-birds,akhtar-etal-2023-exploring}, temporal~\citep{tan-etal-2023-towards}, and causal reasoning~\citep{du-etal-2022-e}.

More recently, research has begun to examine the cultural dimensions of commonsense reasoning, investigating how language models encode and generalize culturally grounded knowledge. \citet{tacl/koto-etal-2024-indoculture} introduced \texttt{IndoCulture}, a dataset for evaluating commonsense reasoning across eleven Indonesian provinces, while \citet{acl/sadallah-etal-2025-commonsense} proposed a complementary benchmark for Arab culture. Other efforts have explored cultural variation beyond commonsense reasoning: \citet{durmustowards} introduced \texttt{GlobalOpinionQA}, built on the World Values Survey~\citep{haerpfer2022world}, to analyze cross-national differences in LLM-generated responses, and \texttt{CultureNLI}~\citep{huang-yang-2023-culturally} examined entailment across Indian and American cultural contexts. In our downstream tasks (Section~\ref{sec:downstream_eval}), we focus on \texttt{IndoCulture} and \texttt{ArabCulture}, as they directly target cultural commonsense reasoning and are manually curated with high-quality annotations.

\section{Cultural Commonsense Knowledge Graph (CCKG)}

In this section, we detail our framework for constructing the Cultural Commonsense Knowledge Graph (CCKG) from LLMs.

\subsection{Preliminaries}

Let 
\[
G = (V, E)
\] 
be a directed labeled graph, where $V$ denotes the set of actions and 
$E \subseteq V \times R \times V$ represents the set of labeled edges. Each edge $(A_i, R, A_j) \in E$ encodes an assertion of the form
\[
A_i \xrightarrow{R} A_j,
\]
interpreted as ``if action $A_i$ occurs, then a related action $A_j$ follows,'' connected through relation $R$.

\noindent\textbf{Actions.} 
Actions are phrases that describe activities, events, or processes representing culturally grounded behaviors.

\noindent\textbf{Assertion.} 
An assertion is a triple $(A_i, R, A_j) \in E$ capturing culturally specific commonsense inferences as a conditional link between an initiating action $A_i$ and a resulting action $A_j$ via relation $R$, which defines their causal, motivational, or consequential connection.

\noindent\textbf{Path.}
A path is an ordered sequence of assertions that connects an initiating action to a resulting action via their relations. 
%Formally, a path in $G$ is a finite sequence of actions and relations:
\[
A_0 \xrightarrow{R_1} A_1 \xrightarrow{R_2} A_2 \xrightarrow{R_3} \dots \xrightarrow{R_k} A_k,
\]
where each $(A_{i-1}, R_i, A_i) \in E$ for $i = 1, \dots, k$, $A_0$ is the initial action, $A_k$ is the resulting or extended action, $A_1, \dots, A_{k-1}$ are intermediate actions, and $R_1, \dots, R_k$ specify the relations connecting them.

\noindent\textbf{Relation.} 
A relation connects two actions and defines the type of inferential link between them.  We categorize five relation types (\texttt{xNext}, \texttt{xEffect}, \texttt{xNeed}, \texttt{oNext}, \texttt{oEffect}), summarized with illustrative examples in Table~\ref{tab:relations}.

\begin{table*}[t]
    \centering
    \resizebox{0.95\linewidth}{!}{%
    \small
    \renewcommand{\arraystretch}{1.2}
    \setlength{\tabcolsep}{8pt}
    \begin{tabular}{l p{4.2cm} p{6.8cm}}
        \toprule
        \textbf{Relation} & \textbf{Definition} & \textbf{Example} \\
        \midrule
        \texttt{xNext} & What would x likely want to do after the action? & 
        buys groceries $\rightarrow$ want to cook (If x buys groceries, x will want to cook.) \\
        
        \texttt{xEffect} & What effects does the action have on x? & 
        gives a gift $\rightarrow$ gets thanked (If $x$ gives a gift, x gets thanked.) \\
        
        \texttt{xNeed} & What does x need to do before the action? & 
        cook a meal $\rightarrow$ gather ingredients (Before x can cook a meal, x needs to gather ingredients.) \\
        
        \texttt{oNext} & What would others likely want to do after the action? & 
        calls the police $\rightarrow$ dispatch officers (If x calls the police, others want to dispatch officers.) \\
        
        \texttt{oEffect} & What effects does the action have on others? & 
        insults someone $\rightarrow$ feel angry (If x insults someone, others will feel angry.) \\
        \bottomrule
    \end{tabular}
    }
    \caption{Types of relations between actions. Three are adopted from \texttt{ATOMIC} \citealt{aaai/atomic/SapBABLRRSC19} (\texttt{oEffect}, \texttt{xEffect}, \texttt{xNeed}) and two are newly introduced (\texttt{oNext}, \texttt{xNext}). In these relation types, x denotes the agent or primary person performing the action, while o refers to others who interact with or are affected by x's action. Text in brackets shows the "if $action_1$, then $action_2$" version.}
    \label{tab:relations} 
\end{table*}

\subsection{Knowledge Graph Construction}
\label{sec:extraction}

\begin{algorithm}[ht]
\caption{CCKG Construction for country $c$ in language $L$}
\label{alg:cckg}
\resizebox{0.85\linewidth}{!}{%
\begin{minipage}{\linewidth}
\KwIn{Country $c$, language $L$, subtopics $\mathcal{S}$, expansion depth $N$}
\KwOut{Knowledge base $\mathcal{K}_c^L$}

\textbf{Init:} $\mathcal{K}_c^L \gets \emptyset$, $\mathcal{U} \gets \emptyset$, $\mathcal{L}_0 \gets \emptyset$ \tcp*{knowledge base, unique actions, expansion list}
\textbf{1) Initial generation} \\
\ForEach{$s \in \mathcal{S}$}{
    Generate assertions $\mathcal{A}_s$ using \texttt{Prompt \ref{fig:initial-prompt}}, insert into $\mathcal{K}_c^L$ \;
    $\mathcal{U} \gets \mathcal{U} \cup \{A_i, A_j \mid (A_i,R,A_j) \in \mathcal{A}_s\}$\; $\mathcal{L}_0 \gets \mathcal{L}_0 \cup \mathcal{A}_s$ \;
}
\textbf{2) Iterative expansion} \\
\For{$t = 1$ \KwTo $N$}{
    $\mathcal{L}_t \gets \emptyset$ \;
    \ForEach{$(A_i, R, A_j) \in \mathcal{L}_{t-1}$ with $R \in \{\texttt{oNext}, \texttt{xNext}\}$}{
        
        \tcp{Intermediate expansion}
        Generate decomposition $A_i \xrightarrow{R_0} A_{i_1} \xrightarrow{R_1} \dots \xrightarrow{R_k} A_j$ using \texttt{Prompt \ref{fig:expansion-prompt}} \;
        Insert intermediate assertions into $\mathcal{K}_c^L$\; 
        Update $\mathcal{U} \gets \mathcal{U} \cup \{A_{i_1},\dots,A_{i_k}\}$ \;
        
        \tcp{Forward expansion}
        Generate new assertions $(A_j, R', A_k)$ with $R' \in \{\texttt{oNext}, \texttt{xNext}\}$ using Prompt \ref{fig:expansion-prompt} \;
        
        \ForEach{$(A_j, R', A_k)$ generated}{
            $u^\star = \arg\max_{u \in \mathcal{U}} \text{sim}(A_k, u)$, $\sigma = \max_{u\in\mathcal{U}}\text{sim}(A_k,u)$ \;
            \eIf{$\sigma > 0.8$}{
                Replace $A_k$ with $u^\star$, insert $(A_j, R', u^\star)$ into $\mathcal{K}_c^L$ \;
            }{
                Insert $(A_j, R', A_k)$ into $\mathcal{K}_c^L$\; 
                Retain at most 6 assertions per $A_j$ in $\mathcal{L}_t$\ for expansion\; Add $(A_j, R', A_k)$ to $\mathcal{L}_t$\tcp*{candidate for next iteration} 
                Update $\mathcal{U} \gets \mathcal{U} \cup \{A_k\}$\;
            }
        }
    }
}
\end{minipage}
}
\end{algorithm}  

We designed an iterative, prompt-based method to generate CCKG. It can be constructed either in English or in the target country's native language. The overall construction procedure of the CCKG is summarized in Algorithm~\ref{alg:cckg} and consists of two stages:
% \begin{enumerate}

\paragraph{Initial Generation.} Given a target language $L$ (English or the native language of the country $c$), we prompt the LLM (using prompt \ref{fig:initial-prompt} in the Appendix) to generate, for each subtopic $s$, possible cultural assertions guided by the predefined relation types. Each assertion is expressed as a triple $(A_i, R, A_j)$, where $A_i$ and $A_j$ are actions and $R$ is one of the relations defined in Table~\ref{tab:relations}. The resulting assertions are stored in the knowledge base $\mathcal{K}_c^L$. 
    
\paragraph{Iterative Expansion.} Starting from assertions $(A_i, R, A_j)$ with $R \in \{\texttt{oNext}, \texttt{xNext}\}$, the knowledge base is enriched over $N$ user-specified number of expansions by elaborating paths. Expansion proceeds in two steps: \textbf{(i) Intermediate action expansion.} The LLM decomposes each $(A_i, R, A_j)$ into a sequence of intermediate steps $A_i \xrightarrow{R_0} A_{i_1} \xrightarrow{R_1} A_{i_2} \xrightarrow{R_2} \cdots \xrightarrow{R_k} A_j$, adding intermediate triples to $\mathcal{K}_c^L$; \textbf{(ii) Forward expansion.}  Using $A_j$ as the new starting point, the LLM generates generates new continuations $(A_j, R', A_k)$ where $R' \in \{\texttt{oNext}, \texttt{xNext}\}$, conditioned on the original context $(A_i, R, A_j)$. To avoid redundant expansions, we maintain a set $\mathcal{U}$ containing all unique actions already present in $\mathcal{K}_c^L$. For each newly generated assertion $(A_j, R', A_k)$, candidate actions $A_k$ are matched against existing actions $\mathcal{U}$ via a similarity score, replacing $A_k$ with $u \in \mathcal{U}$ if $\text{sim}(A_k,u) > 0.8$. Otherwise, $(A_j,R',A_k)$ is inserted as a novel assertion in $\mathcal{K}_c^L$ and included in the pool of candidates for further expansion. Because the number of assertions generated for each $A_j$ may vary, we retain at most six assertions per list for expansion in the following iteration. This pruning strategy balances computational efficiency with knowledge coverage. 
% \end{enumerate}
Both steps of the iterative expansion leverage the prompt \ref{fig:expansion-prompt} in the Appendix to instruct the LLM to produce intermediate actions and forward actions.

\section{Experiments}

\subsection{CCKG Extraction}
\label{sec:evaluation-metrics}

\paragraph{Topic Taxonomy and Country Selection.}
We selected five countries: China, Indonesia, Japan, England, and Egypt, to ensure broad geographic coverage, cultural diversity, and the availability of native human evaluators. Their corresponding native languages are Chinese (CHI), Bahasa Indonesia (IND), Japanese (JAP), English (EN), and Modern Standard Arabic (MSA). We defined 11 daily-life topics comprising 65 fine-grained subtopics, adapted from \citet{tacl/koto-etal-2024-indoculture}. These topics cover diverse aspects of everyday life, including food, weddings, art, habits, daily routines, family relationships, pregnancy and child-rearing, death, religious holidays, traditional games, and socio-religious practices. A full taxonomy of topics and subtopics is presented in Table~\ref{tab:topics_subtopics} in the Appendix. While our current selection of countries is limited to this experimental setup, the proposed framework is general and can be extended to any cultural or linguistic context.

\paragraph{Evaluation Setup.}

To assess the quality of CCKG, we conduct a manual evaluation of assertions and their derived paths across three dimensions using binary labels (yes/no), following prior work~\cite{www/NguyenRVW23-Mango,bhatia-shwartz-2023-gd}. Specifically, we assess:
(i) \textit{Correctness} (\textbf{COR}): whether $A_i$ and $A_j$ are valid actions and the relation $R$ accurately represents their connection (see Table~\ref{tab:relations});
(ii) \textit{Cultural relevance} (\textbf{CR}): whether the assertion is culturally specific to the target country, as opposed to being universal or broadly cross-cultural;
(iii) \textit{Logical path coherence} (\textbf{LPC}): whether a sequence of actions forms a coherent, logically structured, and contradiction-free inferential chain.

The evaluation was conducted by expert annotators who are native speakers of the corresponding languages, possess at least a high-school diploma, and are proficient in both English and their native language (see \S\ref{sec:annot_criteria} in the Appendix for full eligibility criteria). For each country, we recruited two evaluators. To discourage careless responses, each evaluation set included five randomly embedded gold-standard samples for quality control, and annotators were required to correctly label at least four of them. All annotators were compensated at their country's minimum wage, and each task took approximately three hours to complete on average.
%and conducted a pilot test to identify reliable annotators. 

\paragraph{Preliminary Experiments.}
To identify the most suitable model for our main experiments, we compare two strong candidates: GPT-4o~\cite{openai2024gpt4ocard} as a closed-source representative and Llama-3.3-70B-IT~\citep{grattafiori2024llama3herdmodels} 
%(hereafter Llama-3.3-70B) 
as its open-source counterpart. This preliminary study focuses on China and Indonesia, and evaluates only the first-stage extraction (Section~\ref{sec:extraction}), as the quality of the subsequent \emph{iterative expansion} stage critically depends on these initial outputs. We randomly sample 100 assertions across 11 topics for each country and ask native evaluators from the corresponding countries to assess \textbf{CR} and \textbf{COR}.\footnote{\textbf{LPC} is excluded since this experiment involves only the initial extraction stage.} As shown in Figure~\ref{fig:gpt_vs_Llama} (Appendix~\ref{app:sec:model_selection}), GPT-4o consistently outperforms Llama-3.3-70B, and we therefore adopt GPT-4o for all subsequent experiments.

\paragraph{Extraction and Evaluation.}

We apply Algorithm~\ref{alg:cckg} using GPT-4o for each country, generating assertions in both English and the respective native language (temperature = 1, $N=3$). To remove duplicates, we use sentence embeddings (\texttt{all-MiniLM-L6-v2}\footnote{\url{https://www.sbert.net/docs/sentence_transformer/pretrained_models.html}}
 for English and \texttt{stsb-xlm-r-multilingual}\footnote{\url{https://huggingface.co/sentence-transformers/stsb-xlm-r-multilingual}}
 for other languages). After filtering out duplicates and malformed assertions, 37,363 English (of 38,858) and 16,709 native-language (of 17,043) assertions remain. We then construct simple paths for each subtopic by treating every initial action $A_i$ as a source node, resulting in 27,649 English and 6,571 native-language paths. Detailed dataset statistics are provided in Table~\ref{tab:dataset_stats} (Appendix).

\paragraph{Result.}

\begin{table}[t]
\centering
\resizebox{0.8\columnwidth}{!}{%
\begin{tabular}{lccc}
\toprule
\textbf{Country (Language)} & \textbf{CR} & \textbf{COR} & \textbf{LPC} \\
\midrule
England (EN) & \textbf{40.0} & \textbf{96.6} & \textbf{82.9} \\
\hline
China (EN) & \textbf{80.8} & \textbf{86.9} & \textbf{70.2} \\
China (CHI) & 59.1 & 85.2 & 59.9 \\
\hline
Egypt (EN) & \textbf{56.9} & \textbf{89.2} & \textbf{96.1} \\
Egypt (MSA) & 13.4 & 82.8 & 89.9 \\
\hline
Japan (EN) & \textbf{72.7} & \textbf{88.3} & 42.9 \\
Japan (JAP) & 55.9 & {83.4} & \textbf{63.9} \\
\hline
Indonesia (EN) & 40.0 & \textbf{81.0} & 72.2 \\
Indonesia (IND) & \textbf{42.1} & 70.7 & \textbf{73.9} \\
\bottomrule
\end{tabular}
}
\caption{
Average percentage of positive annotations (\textit{yes} labels) for Correctness (COR), Cultural Relevance (CR), and Logical Path Coherence (LPC) across two annotators. Bold values indicate higher scores between English and native-language CCKG for each country and criterion.
}
\label{tab:annotator_agreement}
\end{table}

To assess how language choice affects CCKG quality, we compared graphs generated in English against those produced in the corresponding native languages. As shown in Table~\ref{tab:annotator_agreement}, English CCKG consistently outperform native-language versions across nearly all evaluation dimensions. On average, English versions achieve higher scores in correctness, cultural relevance, and logical path coherence, indicating that LLMs express cultural knowledge more accurately and coherently when operating in English. This pattern holds across diverse linguistic families---including Arabic, Chinese, and Japanese---suggesting that English serves as a more stable representational medium for encoding culturally grounded reasoning. Native-language CCKG, while sometimes capturing localized nuances, tend to produce less coherent or less contextually grounded inferential chains, likely due to limited language-specific pretraining data. Overall, these findings highlight a key asymmetry in current multilingual LLMs: despite aiming to model local cultural reasoning, they still represent cultural commonsense most effectively through English.

\subsection{Evaluation on Cultural Commonsense Reasoning}\label{sec:downstream_eval}

\paragraph{Dataset.} We evaluate whether integrating LLMs with cultural inferential knowledge from CCKG—used as in-context exemplars—enhances their performance on tasks requiring culturally grounded reasoning. We use two human-constructed benchmarks: \texttt{ArabCulture}~\cite{acl/sadallah-etal-2025-commonsense} and \texttt{IndoCulture}~\cite{tacl/koto-etal-2024-indoculture}. \texttt{ArabCulture} covers cultural commonsense from 13 Arab countries (including Egypt) and is written in Modern Standard Arabic (MSA), while \texttt{IndoCulture} represents cultural reasoning across 11 Indonesian provinces in Bahasa Indonesia. Both datasets span diverse cultural domains and everyday life scenarios, and can be evaluated in two formats: (i) multiple-choice question answering (MCQA), where each instance presents three candidate completions with exactly one correct answer, and (ii) sentence completion tasks (i.e., open-ended generation). All evaluations are conducted in both English and the respective native languages of each country.

\paragraph{Models.} We experimented with 13 models in total: base models Llama3.2-1B/3B, Llama3.1-8B \cite{grattafiori2024llama3herdmodels}, Qwen2.5-0.5B/1.5B/3B/7B \cite{qwen2025qwen25technicalreport}, and Gemma2-2B/9B \cite{gemmateam2024gemma2improvingopen}; and instruction-tuned models Llama3.1-8B-I, Gemma2-9B-I, and Qwen2.5-7B-I.
All models are used for cultural commonsense question answering, while only the instruction-tuned models are used for generation tasks, including cultural commonsense completion and story generation.

\paragraph{Augmentation Methods.}

We perform in-context augmentation with relevant assertions (5-shot, \textit{\textbf{\mbox{-Asrt}}}) or paths (1-shot, \textit{\textbf{-Path}}), on both MCQA and sentence completion tasks. Here, we use SBERT embeddings \cite{reimers-2019-sentence-bert}  with \texttt{stsb-xlm-r-multilingual}\footnote{\url{https://huggingface.co/sentence-transformers/stsb-xlm-r-multilingual}} for semantic search\footnote{\url{https://github.com/UKPLab/sentence-transformers/blob/master/examples/applications/semantic-search/semantic_search.py}} to retrieve the most relevant assertions and paths. 

As baselines, we include (i) zero-shot prompting without in-context augmentation, denoted as \textit{\textbf{Base}}, and (ii) chain-of-thought prompting (CoT;~\citealt{10.5555/3600270.3602070}) for MCQA. We also compare with in-context augmentation using \texttt{Mango} (5-shot, \textit{\textbf{Mango}}), a widely used LLM-extracted cultural commonsense knowledge base that provides factual assertions but does not model paths.

\begin{table*}[t]
\centering
% \small
\resizebox{\textwidth}{!}{%
\setlength{\tabcolsep}{0.7mm}
\begin{tabular}{lrrrrrrrcrrrrrrr}
\toprule
& \multicolumn{7}{c}{\cellcolor{blue!7} \textbf{IndoCulture}} & & \multicolumn{7}{c}{\cellcolor{red!7}\textbf{ArabCulture}} \\
\cmidrule(lr){2-8}\cmidrule(lr){9-16}
\multirow{2}{*}{\textbf{Models}} 
  & \multicolumn{2}{c}{\cellcolor{blue!7}\textbf{Before Aug}} 
  & \multicolumn{5}{c}{\cellcolor{blue!7}\textbf{After Aug}} 
  &
  & \multicolumn{2}{c}{\cellcolor{red!7}\textbf{Before Aug}} 
  & \multicolumn{5}{c}{\cellcolor{red!7}\textbf{After Aug}} \\
\cmidrule(lr){2-3} \cmidrule(lr){4-8}
\cmidrule(lr){10-11} \cmidrule(lr){12-16}
& \cellcolor{blue!7}\textbf{Base} & \cellcolor{blue!7}\textbf{CoT} 
& \cellcolor{blue!7}\textbf{Mango} 
& \cellcolor{blue!7}\textbf{E-Asrt} & \cellcolor{blue!7}\textbf{E-Path} & \cellcolor{blue!7}\textbf{N-Asrt} & \cellcolor{blue!7}\textbf{N-Path}
&
& \cellcolor{red!7}\textbf{Base} & \cellcolor{red!7}\textbf{CoT} 
& \cellcolor{red!7}\textbf{Mango} 
& \cellcolor{red!7}\textbf{E-Asrt} & \cellcolor{red!7}\textbf{E-Path} & \cellcolor{red!7}\textbf{N-Asrt} & \cellcolor{red!7}\textbf{N-Path} \\
\midrule
Qwen2.5-0.5B     & 43.1 & 35.8 & \textbf{43.5} & 43.4 & 42.2 & 41.6 & 41.9 
& & 33.3 & {34.2} & \textbf{34.3} & 33.8 & 33.8 & 34.2 & 34.2 \\
Qwen2.5-1.5B     & 43.8 & \textbf{45.5} & 45.0 & 45.2 & 44.8 & 44.5 & 41.8 
& & 39.4 & 43.4 & 43.5 & 45.2 & 41.4 & \textbf{48.3} & 42.7 \\
Qwen2.5-3B       & 53.0 & 53.1 & \textbf{54.3} & 53.2 & 51.7 & 53.6 & 51.8 & &  37.5 & 35.2 & 40.6 & 44.6 & 39.8 & \textbf{45.7} & 38.1 \\
Qwen2.5-7B       & 58.5 & 53.4 & 59.7 & 60.1 & 59.1 & \textbf{60.8} & 58.6 & & 49.3 & 40.6 & 52.6 & 53.6 & 46.5 & \textbf{59.6} & 51.1 \\
Gemma2-2B         & 33.4 & 35.5 & 38.3 & 37.5 & 38.2 & 35.7 & \textbf{39.4} & & 34.3 & 34.3 & 34.3 & 34.2 & 34.0 & \textbf{34.6} & 34.0 \\
Gemma2-9B         & 65.2 & 53.2 & 64.8 & 64.6 & 65.0 & 65.7 & \textbf{65.9} & & \textbf{34.5} & 34.3 & 34.3 & 34.3 & 34.3 & 34.3 & 34.3 \\
Llama3.2-1B       & 46.9 & 44.5 & 48.9 & 50.9 & 51.0 & \textbf{53.4} & 53.1 & & \textbf{33.9} & 33.8 & 33.7 & 33.8 & 33.8 & 33.8 & 33.4 \\
Llama3.2-3B       & 49.4 & 41.3 & 49.2 & 49.6 & 49.4 & 50.0 & \textbf{50.5} & & 33.3 & \textbf{37.3} & 34.0 & 34.0 & 33.0 & 34.4 & 31.9 \\
Llama3.1-8B       & 32.7 & \textbf{35.6} & 32.7 & 32.7 & 32.7 & 32.7 & 32.7 & & 34.9 & 35.3 & 34.5 & 34.8 & 34.2 & \textbf{35.4} & 34.7 \\
\midrule
Gemma2-9B-IT      & 57.9 & 39.1 & 57.9 & 57.3 & 59.1 & 59.4 & \textbf{61.0} & & \textbf{57.3} & 34.3 & 47.0 & 43.7 & 46.8 & 42.9 & 49.3 \\
Qwen2.5-7B-IT     & 66.2 & 63.5 & 65.3 & 66.1 & 66.9 & 66.3 & \textbf{67.5} & & 48.7 & 34.2 & 46.5 & 47.8 & 49.3 & \textbf{50.8} & 49.2 \\
Llama3.1-8B-IT     & 55.5 & \textbf{59.0} & 53.4 & 54.2 & 54.4 & 54.4 & 56.4 & & \textbf{49.3} & 34.4 & 35.4 & 37.0 & 40.1 & 37.2 & 39.3 \\
\midrule
{Avg $\Delta$} & NA & $-$3.8 & 0.6 & 0.8 & 0.7 & 1.0 & \textbf{1.2} & & NA & $-$4.6 & $-$1.3 & $-$0.7 & $-$1.5 & \textbf{0.4} & $-$1.2 \\
\bottomrule
\end{tabular}
}
\caption{Accuracy comparison on MCQA across different methods. \textbf{E-Asrt}: English CCKG Assertions; \textbf{E-Path}: English CCKG Paths; \textbf{N-Asrt}: Native-language (Arabic or Indonesian) CCKG Assertions; \textbf{N-Path}: Native-language (Arabic or Indonesian) CCKG Paths. ``Avg $\Delta$'' denotes the average improvement over the baseline. Best results per model are highlighted in bold. }
\label{tab:results_english_mcq}
\end{table*}

\paragraph{Evaluation Metrics.}
For MCQA, we report accuracy using the official evaluation scripts and generation parameters provided by each benchmark. For sentence completion, we evaluate using BERTScore-F1~\cite{Zhang*2020BERTScore:} and sentence similarity~\cite{corley-mihalcea-2005-measuring}, computed between the LLM-generated text and the corresponding reference completion. All experiments use the original benchmark prompts.

\paragraph{Results on MCQA.}

Table~\ref{tab:results_english_mcq} presents the MCQA accuracy across models and augmentation methods using English prompts, while results for Arabic and Indonesian prompts are provided in Appendix~\ref{sec:Mcqa_native} and~\ref{sec:sentence_completion_native}. Overall, integrating CCKG knowledge—either through assertions or paths—consistently improves performance in \texttt{IndoCulture}. For \texttt{ArabCulture}, the improvement is not observed in Mango, but only in our CCKG assertion. The largest gains are observed when models are augmented with native-language assertions, suggesting that culturally grounded examples expressed in the original language are more effective in guiding model predictions. For instance, Qwen2.5-7B improves from 58.5\% to 60.8\% on \texttt{IndoCulture} and from 49.3\% to 59.6\% on \texttt{ArabCulture} when augmented with Indonesian and Arabic assertions, respectively. On average, native-language augmentation achieves the highest improvement ($+$1.2 points), outperforming English-based augmentation and Mango.

Smaller or base models benefit the most from in-context augmentation, implying that explicit inferential cues from CCKG compensate for their limited internalized cultural knowledge. Larger instruction-tuned models, such as Gemma2-9B-IT and Qwen2.5-7B-IT, show more modest or mixed gains, likely because they already encode general cultural knowledge, making additional context less impactful.

In contrast, chain-of-thought prompting (CoT) performs worse than the baseline across nearly all models and datasets, with average drops of 3.8 points on \texttt{IndoCulture} and 4.6 points on \texttt{ArabCulture}. This suggests that cultural commonsense reasoning relies more on intuitive and context-dependent knowledge than on step-by-step logical reasoning—a finding consistent with prior observations that explicit reasoning often weakens culturally situated inference~\cite{acl/sadallah-etal-2025-commonsense}.\footnote{Similar trends are also observed with native prompts (\S\ref{sec:Mcqa_native}).}

\paragraph{Results on Sentence Completion.}
As shown in Table~\ref{tab:similarity_bert_results_completion}, incorporating context from CCKG generally improves both similarity and BERTScore-F1. Native-language assertions and paths yield the highest gains in similarity, with modest but consistent improvements in BERTScore, outperforming both the baseline without augmentation and Mango augmentation.  
For example, compared to the baseline, Llama3.1-8B-IT improves similarity scores from 32.6\% to 36.0\% on IndoCulture and from 29.9\% to 33.5\% on ArabCulture. Qwen2.5-7B-IT shows comparable gains, increasing from 39.6\% to 43.0\% and from 38.8\% to 42.4\%, respectively. BERTScore changes are modest, with Llama3.1-8B-IT exhibiting slight drops (65.0\% to 64.5\% on ArabCulture, 70.2\% to 70.1\% on IndoCulture), while Qwen2.5-7B-IT shows marginal improvements (72.3\% to 72.6\% on ArabCulture) compared to the baseline. Similar trends are observed with prompts in native language (see \S\ref{sec:sentence_completion_native}).

\begin{table}[t]
\centering
% \footnotesize
\resizebox{\columnwidth}{!}{%
\setlength{\tabcolsep}{1.8pt}
\begin{tabular}{lccccccc}
\toprule
\multirow{2}{*}{\textbf{Models}} & \multirow{2}{*}{\begin{tabular}[c]{@{}c@{}}\textbf{Before}\\ \textbf{Aug}\end{tabular}} & \multirow{2}{*}{\textbf{+Mango}} & \multicolumn{4}{c}{\textbf{+CCKG Methods}} \\
\cmidrule(lr){4-7}
& & & \textbf{E-Asrt} & \textbf{E-Path} & \textbf{N-Asrt} & \textbf{N-Path} \\
\midrule
\multicolumn{7}{c}{\cellcolor{blue!7} \textbf{IndoCulture w/ Sentence Similarity Score}} \\
\midrule
Gemma2-9B-IT & 32.0 & 33.3 & 34.4 & 34.6 & \textbf{35.5} & 35.4 \\
Qwen2.5-7B-IT & 39.6 & 41.7 & 42.5 & 42.6 & 42.5 & \textbf{43.0} \\
Llama3.1-8B-IT & 32.6 & 33.8 & 34.3 & 34.9 & 36.1 & \textbf{36.0} \\
\textbf{Avg} & 34.8 & 36.3 & 37.1 & 37.3 & 38.0 & \textbf{38.1} \\
\midrule
\multicolumn{7}{c}{\cellcolor{blue!7} \textbf{IndoCulture w/ Avg BERT Score F1}} \\
\midrule
Gemma2-9B-IT & 71.2 & 71.0 & 71.3 & 71.3 & \textbf{71.5} & 71.4 \\
Qwen2.5-7B-IT & 72.3 & 72.3 & \textbf{72.6} & 72.5 & \textbf{72.6} & 72.5 \\
Llama3.1-8B-IT & \textbf{70.2} & 69.6 & 69.9 & 70.0 & 70.1 & 70.0 \\
\textbf{Avg} & 71.3 & 71.0 & 71.3 & 71.2 & \textbf{71.4} & 71.3 \\
\midrule
\multicolumn{7}{c}{\cellcolor{red!7} \textbf{ArabCulture w/ Sentence Similarity Score}} \\
\midrule
Gemma2-9B-IT & 33.9 & 33.0 & 36.5 & 32.9 & \textbf{37.2} & 34.1 \\
Qwen2.5-7B-IT & 38.8 & 39.2 & 42.0 & 35.3 & \textbf{42.5} & 37.6 \\
Llama3.1-8B-IT & 29.9 & 29.1 & 31.2 & 26.6 & \textbf{33.5} & 29.4 \\
\textbf{Avg} & 34.2 & 33.8 & 36.5 & 31.6 & \textbf{37.7} & 33.7 \\
\midrule
\multicolumn{7}{c}{\cellcolor{red!7} \textbf{ArabCulture w/ Avg BERT Score F1}} \\
\midrule
Gemma2-9B-IT & 68.6 & 68.4 & 68.7 & 68.6 & \textbf{68.7} & 68.6 \\
Qwen2.5-7B-IT & 67.8 & 67.4 & 67.4 & 65.7 & \textbf{67.8} & 66.5 \\
Llama3.1-8B-IT & \textbf{65.0 }& 63.5 & 63.0 & 63.2 & 64.5 & 64.0 \\
\textbf{Avg} & 67.0 & 66.4 & 66.3 & 65.8 & \textbf{67.0} & 66.4 \\
\bottomrule
\end{tabular}
}
\caption{Sentence similarity and BERT scores for sentence completion task. 
\textbf{E-Asrt}: CCKG English Assertions, \textbf{E-Path}: CCKG English Paths, \textbf{N-Asrt}: CCKG Native-language Assertions, \textbf{N-Path}: CCKG Native-language Paths.
Best results per row are bolded.}
\label{tab:similarity_bert_results_completion}
\end{table}

\subsection{Evaluation on Story Generation}
To further examine the usefulness of CCKG paths, we conducted a free-form short story generation task covering 25 randomly selected subtopics (see prompts in \S\ref{sec:story_gen_prompt}). Stories were generated in both English and the native languages of Egypt, China, and Indonesia—chosen based on the availability of qualified human evaluators. We compared three setups: baseline zero-shot prompting, in-context inference with \emph{+Mango} (5-shot assertions from Mango), and \emph{+CCKG} (1-shot paths retrieved from CCKG). Relevant assertions or paths were selected using SBERT embeddings, following the same retrieval procedure described in \S\ref{sec:downstream_eval}. For CCKG, we focus on path-based augmentation here, as story generation naturally benefits from sequential and causal structure.

\paragraph{Evaluation Metrics.}
For the evaluation, we primarily relied on human judgments. Two annotators rated each story on a 1--10 Likert scale along three dimensions: 1) \emph{Cultural relevance} (\textbf{CR}), which measures how accurately the story reflects the traditions, customs, values, and social norms of the country; 2) \emph{Fluency} (\textbf{FL}), which assesses grammatical correctness, sentence structure, vocabulary, and readability; and 3) \emph{Coherence} (\textbf{CO}), which analyzes the logical flow, clarity, and consistency of events and character actions. As a complementary analysis, we also employed LLM-as-a-Judge \cite{qiu-etal-2025-evaluating,li2024llmsasjudgescomprehensivesurveyllmbased}, using GPT-4o (prompts in Appendix~\ref{sec:story_gen_prompt}) with the same evaluation criteria and examining its correlation with human judgments.

\paragraph{Results.} Table~\ref{tab:results_story_english} summarizes the aggregated human evaluation scores for English story generation. Across all nine model–country pairs, incorporating CCKG paths consistently improves story quality, with the largest gains observed in \textit{Cultural Relevance}—averaging a +1.4 increase for Llama models over the baseline across three countries. \textit{Fluency} and \textit{Coherence} also show steady, smaller improvements, suggesting that path-based augmentation helps models produce more logically structured and contextually grounded narratives. Results for native-language story generation show greater variation and are detailed in Appendix~\ref{sec:native_story_gen_result}.

\begin{table}[t]
\centering
\resizebox{\columnwidth}{!}{%
\small
\setlength{\tabcolsep}{3pt}
\begin{tabular}{@{}lccccccccc@{}}
\toprule
& \multicolumn{3}{c}{\textbf{China}} & \multicolumn{3}{c}{\textbf{Indonesia}} & \multicolumn{3}{c}{\textbf{Egypt}} \\
\cmidrule(lr){2-4} \cmidrule(lr){5-7} \cmidrule(lr){8-10}
& \bf CR & \bf FL & \bf CO & \bf CR & \bf FL & \bf CO & \bf CR & \bf FL & \bf CO \\
\midrule
Llama3.1-8B-IT & 6.3 & 8.4 & 7.6 & 6.9 & 8.1 & 7.7 & 6.3 & 8.9 & 8.7 \\
+Mango & 7.0 & 8.4 & 7.6 & 7.0 & 8.3 & 8.0 & 7.0 & 9.1 & 8.9 \\
+CCKG & \textbf{7.3} & \textbf{9.0} & \textbf{8.2} & \textbf{7.7} & \textbf{8.4} & \textbf{8.3} & \textbf{8.7} & \textbf{9.1} & \textbf{9.5} \\
\midrule
Qwen2.5-7B-IT & 7.0 & 8.7 & 7.9 & 6.6 & 7.5 & 7.2 & 7.0 & 8.9 & 8.8 \\
+Mango & 6.9 & 8.9 & 7.9 & 6.8 & 8.3 & 7.7 & 7.3 & 8.9 & 8.8 \\
+CCKG& \textbf{7.3} & \textbf{8.9} & \textbf{8.5} & \textbf{7.3} & \textbf{8.3} & \textbf{7.9} & \textbf{8.9} & \textbf{9.0} & \textbf{9.2} \\
\midrule
Gemma2-9B-IT & 6.5 & 8.8 & 7.7 & 7.6 & 8.7 & 8.4 & 6.5 & 8.5 & 8.4 \\
+Mango & 6.9 & 9.0 & 7.8 & 6.9 & 8.5 & 8.1 & 6.8 & 8.6 & 8.6 \\
+CCKG & \textbf{7.8} & \textbf{9.2} & \textbf{8.7} & \textbf{8.0} & \textbf{8.9} & \textbf{8.8} & \textbf{7.8} & \textbf{8.8} & \textbf{8.7} \\
\bottomrule
\end{tabular} }
\caption{Aggregated annotator scores for English story generation, comparing Base, \textbf{+Mango}, and \textbf{+CCKG}. \textbf{CR}: Cultural relevance, \textbf{FL}: Fluency, \textbf{CO}: Coherence. Best scores are bolded; inter-annotator correlation is strongest for CR (0.72), moderate for CO (0.34), and weak for FL (0.26) (Appendix~\ref{sec:pearson_correlation}).}

\label{tab:results_story_english}
\end{table}

\begin{figure}
    \centering
    \includegraphics[width=\linewidth]{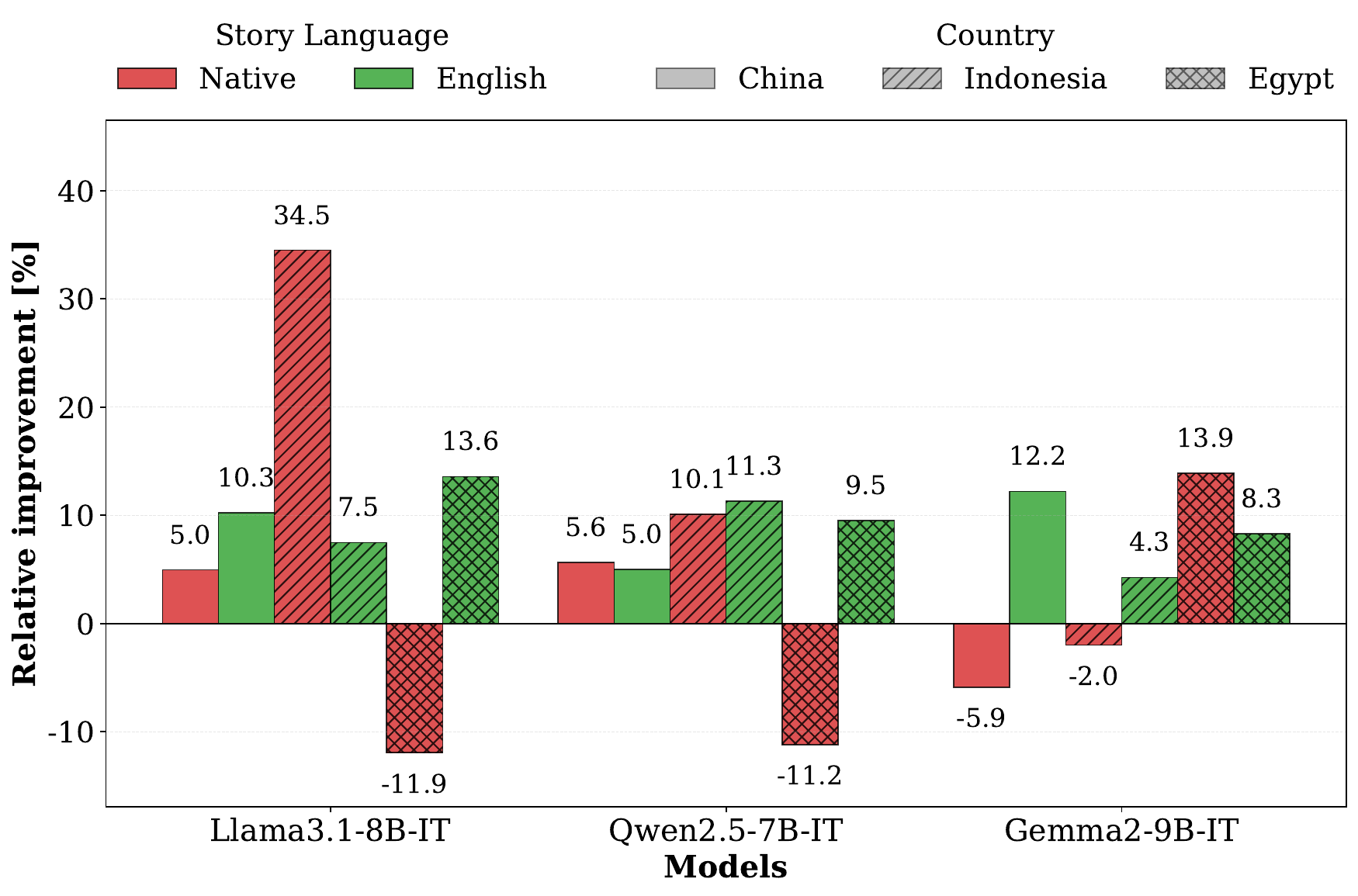}
    \caption{Relative improvement from \textbf{+CCKG} over the baseline in \textit{Native} vs.\ \textit{English} story generation. Bars show percentage lift in average human scores (Cultural relevance, Fluency, Coherence); numbers above bars indicate gains in percentage points.}
    \label{fig:story_eng_vs_nativ}
\end{figure}

Figure \ref{fig:story_eng_vs_nativ} further shows the relative improvements in story quality when augmenting with CCKG over the baseline, across three evaluation metrics for each country and each model. Overall, the benefits of CCKG paths are more evident in English story generation, with Indonesian story generation using Llama3.1-8B-IT being a notable exception. Full per-metric results are provided in Appendix \ref{sec:english_native_wise}. 

\paragraph{LLM-as-a-Judge aligns with human ratings on cultural relevance in native languages but only moderately in English.}
We observe moderate correlations between the LLM judge and human evaluators on \textit{cultural relevance} (average 0.4) in English story generation, but stronger correlations (average 0.8) in native languages. Interestingly, negative correlations are found for English generations in \textit{Fluency} (–0.1) and \textit{Coherence} (0.0), but strong positive correlations for generations in native languages (0.9 and 0.8, respectively). 
One possible explanation is that English evaluations may be more sensitive to stylistic variation—for example, differing preferences for concise versus elaborative writing. In addition, culturally specific or tradition-related expressions often sound more natural and authentic in their native languages than in English translations, which may further influence evaluators’ preferences.
Full LLM evaluation results and correlations with human judgments are provided in Section \ref{sec:llm_as_judge} in the Appendix.

\section{Conclusion}
This paper explores LLMs as cultural technologies and knowledge extractors through \textbf{CCKG}, a framework for constructing multilingual cultural commonsense knowledge chains that extend inferential reasoning beyond static, English-centric resources. By modeling culture as procedural and sequential rather than as isolated facts, \textbf{CCKG} captures the flow of cultural practices across languages. Human evaluations show that while native languages convey richer cultural depth, English outputs are generally more coherent and preferred. Empirically, augmenting LLMs with \textbf{CCKG} improves performance on cultural commonsense reasoning and story generation.

\section*{Limitations}

Our method for constructing the CCKG relies on prompting, which makes it sensitive to the specific prompt formulations. Consequently, some degree of prompt tuning may be required when applying the approach to new models. Nevertheless, we ran experiments with two state-of-the-art language models: a closed-source model (GPT-4o) and an open-source model (Llama-3.3-70B-Instruct). We successfully extracted CCKG from both, demonstrating that our method is robust across different model types.

Automatically extracting cultural commonsense knowledge from LLMs carries the potential risk of reproducing stereotypes. In this work, we did not focus on detecting or evaluating such biases. However, our human evaluators reviewed a subset of the extracted content for quality analysis (see Section \ref{sec:stereo} in Appendix), and the majority of the items were not flagged as stereotypical or harmful material. We plan to conduct a more in-depth investigation of this issue in future work.

In this work, we focus on a limited set of high(er)-resource cultures, reflecting both the accessibility to human evaluators and the assumption that LLMs have already acquired substantial knowledge about these cultures during pre-training. We further evaluate culture at the country level, which we plan on extending to more fine-grained levels in the future. 

We use data extracted from LLMs as a research prototype and as an exploratory foundation for the concept of LLMs as Cultural Archives. Importantly, this extracted content should not be viewed as a formal dataset. We advise against its use in production systems without careful consideration of both the potential benefits—such as enabling more culturally aware technologies—and the corresponding challenges and risks, including the possible reinforcement of stereotypes or other unintended biases.

\section*{Acknowledgments}
We thank the supercomputing center at MBZUAI for providing resources for this research. We also gratefully acknowledge MBZUAI's support in funding the hiring of human annotators. This work was further supported by the LOEWE Distinguished Chair ``Ubiquitous Knowledge Processing'', LOEWE initiative, Hesse, Germany (Grant Number: LOEWE/4a//519/05/00.002(0002)/81).

% Bibliography entries for the entire Anthology, followed by custom entries
%\bibliography{anthology,custom}
% Custom bibliography entries only
\bibliography{custom}

\appendix

\section{Annotator Criteria}
\label{sec:annot_criteria}
We recruited two annotators per country (ten in total) and applied strict eligibility requirements to ensure cultural authenticity and linguistic proficiency. Annotators were required to meet the following criteria:
\begin{itemize}
    \item Native speaker of the specified local language and proficient in English (speaking and comprehension).
\item Resided in the country for at least ten years.
\item Demonstrated deep familiarity with the country's culture.
\item Both parents are also natives residing in the same country.
\item Minimum qualification of senior high school graduation (higher degrees preferred).

\end{itemize}
Among the ten annotators, four held a Bachelor's degree, three a Master's degree, two a Ph.D., and one a postdoctoral qualification. To discourage careless responses,  five gold-standard samples were randomly embedded in each evaluation set for quality control, and annotators were required to correctly label at least four of them. Annotators were compensated at their country's minimum wage, and the task took approximately three hours on average.

%A pilot test was conducted to select reliable annotators who thoroughly understood the guidelines. 
\section{Topic Diversity}
For our study, we defined 11 daily-life topics encompassing 65 fine-grained subtopics (see Table~\ref{tab:topics_subtopics}). Each subtopic was translated into the native languages of the 4 target countries (except England) by native speakers, enabling both English and native-language CCKG generation settings. For Japan and China, additional care was taken during translation to preserve culturally specific nuances.
\begin{table*}[htbp]
\centering 
\normalsize 
\begin{tabular}{p{2.7cm} p{10cm}}
\toprule
\textbf{Topics} & \textbf{Subtopics} \\
\midrule
Food & Breakfast, lunch, dinner, traditional foods and beverages, cutlery, cooking ware, fruit, food souvenirs, snacks \\
\midrule
Wedding & Wedding location, wedding food, wedding dowry, traditions before marriage, traditions when getting married, traditions after marriage, men's wedding clothes, women's wedding clothes, songs and activities during the wedding, invited guests at a wedding, gift brought to weddings, food at a wedding \\
\midrule
Habits & Eating habit, greetings habits, financial habits (saving, debit/credit), punctuality habit, cleanliness habit, shower time habit, transportation habit, popular sports \\
\midrule
Art & Musical instruments, folks songs, traditional dances, use of art at certain events, poetry or similar literature \\
\midrule
Daily activities & Morning activities, afternoon activities, evening activities, leisure and relaxation activities, household activities (cleaning, home management) \\
\midrule
Family relationship & Relationships within the main family, relationships in the extended family, relations with society/neighbors, clan/descendant system \\
\midrule
Pregnancy and kids & Traditions during pregnancy, traditions after birth, how to care for a newborn baby, how to care for toddlers, how to care for teenagers, parents and children interactions as adults \\
\midrule
Death & When death occurs, the process of dealing with a corpse, traditions after the body is buried, the clothes of the mourners, inheritance matters \\
\midrule
Religious holiday & Traditions before religious holidays, traditions leading up to religious holidays, traditions during religious holidays, traditions after holidays \\
\midrule
Traditional games & Traditional game types \\
\midrule
Socio-religious aspects of life & Regular religious activities, mystical things, traditional ceremonies, lifestyle, self care, traditional medicine, traditional sayings \\
\bottomrule
\end{tabular}
\caption{Overview of topics and their associated subtopics.}
\label{tab:topics_subtopics}
\end{table*}

\subsection{Data Statistics}

Table \ref{tab:dataset_stats} shows the dataset statistics of CCKG.

\begin{table}[htbp]
\centering
\resizebox{\columnwidth}{!}{%
\begin{tabular}{lrrrrr}
\toprule
\textbf{Country} & \textbf{Unique} & \textbf{Unique} & \textbf{Total} & \textbf{Avg Path} & \textbf{Eval} \\
\textbf{(Language)} & \textbf{Nodes} & \textbf{Paths} & \textbf{Assertions} & \textbf{Length} & \textbf{Assertions} \\
\midrule
England (EN) & 7698 & 6174 & 8693 & 11.47 & 396 \\
\midrule
Indonesia (EN) & 6267 & 3877 & 6905 & 10.72 & 355 \\
Indonesia (IND) & 5946 & 2398 & 6300 & 7.35 & 220 \\
\midrule
China (EN) & 6082 & 3923 & 6721 & 9.58 & 335 \\
China (CHI) & 3051 & 1179 & 3059 & 5.48 & 297 \\
\midrule
Japan (EN) & 6713 & 7581 & 7565 & 18.23 & 451 \\
Japan (JAP) & 2663 & 1057 & 2629 & 4.07 & 220 \\
\midrule
Egypt (EN) & 6601 & 6094 & 7479 & 17.40 & 393 \\
Egypt (MSA) & 4527 & 1937 & 4721 & 6.33 & 276 \\
\bottomrule
\end{tabular}
}
\caption{Dataset statistics across countries and languages. The Eval Assertions column shows the number of assertion samples to evaluate for 50 paths. The number of edges is the same as the number of assertions.}
\label{tab:dataset_stats}
\end{table}

\section{Software}

In this paper,  we use the Huggingface Transformers library for experiments with cultural commonsense QA. For all the free-form generation tasks, we use the APIs provided by OpenRouter\footnote{https://openrouter.ai}.

\section{Additional Results}

\subsection{Model Selection} \label{app:sec:model_selection}

\begin{figure}
    \centering
    \includegraphics[width=\linewidth]{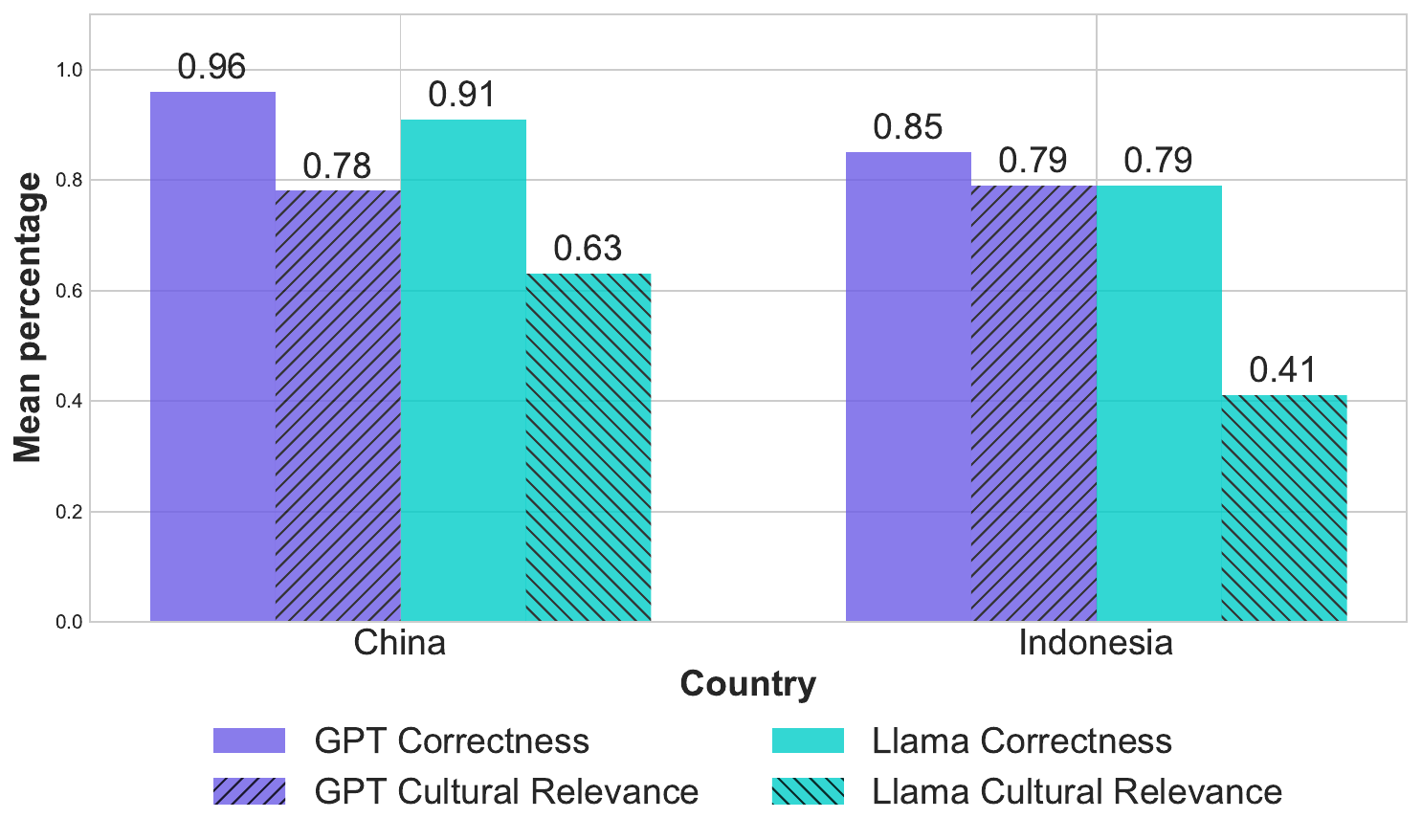}
    \caption{Performance of Llama3.3-70B-IT and GPT-4o on initial generation data to determine the optimal model for CCKG generation.}
    \label{fig:gpt_vs_Llama}
\end{figure}

Figure \ref{fig:gpt_vs_Llama} presents the evaluation results of Llama-3.3-70B-instruct and GPT-4o models on the initial CCKG generation. Overall, the GPT-4o model shows better generation quality measured by cultural relevance and correctness.

\subsection{Native Story Generation Results}
\label{sec:native_story_gen_result}

In native-language story generation, results vary more substantially, reflecting the interaction between language and model alignment (see Table~\ref{tab:results_native_story}). CCKG provides the largest benefit when baseline cultural relevance is weaker---e.g., Llama/Indonesia +1.6 CR ($4.7 \rightarrow 6.3$), Qwen/Indonesia +0.7 ($5.8 \rightarrow 6.5$), and Gemma/Egypt +0.7 ($4.7 \rightarrow 5.4$)---but has little or even negative effect where performance is already strong or misaligned (e.g., Llama/Egypt $-0.0$, Qwen/Egypt $-0.2$). Qwen's Chinese baseline is already high (7.68 cultural relevance) but still benefits from CCKG ($\rightarrow 8.5$, $+0.8$). Fluency and coherence remain largely stable, though dips appear in some cases (e.g., Gemma/Indonesia fluency $8.3 \rightarrow 7.4$), underscoring the importance of language--model fit.

\begin{table}[t]
\centering
\small
\setlength{\tabcolsep}{3pt}
\begin{tabular}{@{}lccccccccc@{}}
\toprule
& \multicolumn{3}{c}{\textbf{China}} & \multicolumn{3}{c}{\textbf{Indonesia}} & \multicolumn{3}{c}{\textbf{Egypt}} \\
\cmidrule(lr){2-4} \cmidrule(lr){5-7} \cmidrule(lr){8-10}
\textbf{Model} & CR & Fl & CO & CR & Fl & CO & CR & Fl & CO \\
\midrule 
%\multicolumn{10}{c}{\textit{Llama3.1-8B-IT}} \\
Llama3.1-8B-IT & 3.6 & 3.8 & 3.4 & 4.7 & 4.9 & 4.6 & 1.9 & 2.1 & 1.9 \\
+Mango & 3.5 & 3.2 & 3.0 & 5.4 & 5.9 & 5.5 & 2.0 & 1.8 & 1.7 \\
+CCKG & \textbf{3.9} & \textbf{3.7} & \textbf{3.8} & \textbf{6.3} & \textbf{6.5} & \textbf{6.4} & 1.9 & 1.6 & 1.7 \\
\midrule
%\multicolumn{10}{c}{\textit{Qwen2.5-7B-IT}} \\
Qwen2.5-7B-IT & 7.7 & 6.8 & 8.2 & 5.8 & 6.2 & 6.1 & 3.0 & 3.0 & 3.3 \\
+Mango & 8.5 & 7.5 & 8.1 & 5.4 & 6.4 & 6.2 & 2.6 & 2.4 & 2.5 \\
+CCKG & \textbf{8.5} & \textbf{7.5} & 7.9 & \textbf{6.5} & \textbf{6.9} & \textbf{6.6} & \textbf{2.8} & \textbf{2.6} & \textbf{2.8} \\
\midrule
%\multicolumn{10}{c}{\textit{Gemma2-9B-IT}} \\
Gemma2-9B-IT & 7.4 & 7.4 & 7.9 & 6.6 & 8.3 & 7.4 & 4.7 & 5.1 & 5.1 \\
+Mango & 7.7 & 7.4 & 7.7 & 6.1 & 7.2 & 6.4 & 4.9 & 5.2 & 5.6 \\
+CCKG & 7.4 & 6.8 & 7.2 & \textbf{7.2} & 7.4 & \textbf{7.3} & \textbf{5.4} & \textbf{5.4} & \textbf{6.1} \\
\bottomrule
\end{tabular}
\caption{Aggregated annotator scores across models and countries for native story generation, comparing the base setting (no augmentation) with two augmented settings: Mango and CCKG. Best scores are bolded.}
\label{tab:results_native_story}
\end{table}

\subsection{Pearson Correlation Between Annotators in Story Generation}
\label{sec:pearson_correlation}
Table \ref{tab:inter-annotator-agreement} reports Pearson correlation scores between the two human annotators, broken down by country, story generation setting, and evaluation criterion. We observe strong agreement in native-language evaluations across all three dimensions—cultural relevance, fluency, and coherence (e.g., Indonesian: 0.9 / 0.8 / 0.7; Chinese: 0.9 / 0.9 / 0.9; MSA: 0.9 / 0.9 / 0.9). In contrast, agreement in English is more variable: correlations are moderate for cultural relevance but substantially weaker for fluency and coherence in the Indonesian and Egyptian sets (e.g., Indonesia-EN: 0.2; Egypt-EN fluency: 0.0), while Chinese-English shows moderate consistency (0.6–0.9). This discrepancy suggests that evaluating stories in English introduces greater variability for fluency and coherence. Two factors likely contribute. First, differences in dialect and stylistic preferences may shape judgments: one annotator may prefer concise, direct English, while another favors more elaborate phrasing or richer descriptive detail. Second, culturally specific expressions and tradition-related terms often sound more natural and authentic in their native languages than in English translation. When such concepts are rendered in English, they may lose nuance or appear less idiomatic, leading to divergent impressions of fluency or coherence.

\begin{table}[t]
\centering
\small
\begin{tabular}{@{}lccc@{}}
\toprule
\textbf{Story} & \textbf{Cultural} & \textbf{Fluency} & \textbf{Coherence} \\
\textbf{Generation} & \textbf{Relevance} & & \\
\midrule
\multicolumn{4}{@{}l}{\cellcolor{blue!7}\textit{Indonesia}} \\
\addlinespace[3pt]
Indonesian & 0.9 & 0.8 & 0.7 \\
English    & 0.4 & 0.2 & 0.2 \\
\midrule
\multicolumn{4}{@{}l}{\cellcolor{blue!7}\textit{China}} \\
\addlinespace[3pt]
Chinese & 0.9 & 0.9 & 0.9 \\
English & 0.9 & 0.6 & 0.7 \\
\midrule
\multicolumn{4}{@{}l}{\cellcolor{blue!7}\textit{Egypt}} \\
\addlinespace[3pt]
MSA     & 0.9 & 0.9 & 0.9 \\
English & 0.9 & 0.0 & 0.2 \\
\bottomrule
\end{tabular}
\caption{Pearson correlation coefficients between Annotator~1 and Annotator~2 for the story-generation task across English and native-language settings.}
\label{tab:inter-annotator-agreement}
\end{table}

\subsection{LLMs-as-Judge for Story Generation Evaluation}
\label{sec:llm_as_judge}

Table \ref{tab:llm-human-agreement} reports Pearson correlations between GPT-4o scores and human judgments. GPT-4o aligns well with human ratings when inter-annotator agreement is high, showing strong correlations in native languages (Indonesian: 0.7 / 0.8 / 0.7; Chinese: 0.7 / 0.8 / 0.7; MSA: 0.9 / 0.9 / 0.9). In contrast, correlations in English are moderate or even negative (e.g., Egypt--EN fluency: $-0.6$; Indonesia--EN coherence: $0.0$).These unstable correlations likely reflect the same factors underlying lower inter-annotator agreement in English (see §\ref{sec:pearson_correlation})—notably differences in stylistic expectations and the fact that culturally specific or tradition-related expressions often sound more natural in their native languages than in English. When such expressions are translated or adapted into English, they may lose nuance or feel less idiomatic, introducing greater variability in perceived fluency and coherence and making English evaluations overall less consistent. Table \ref{tab:results_native_llm_story} and \ref{tab:results_english_llm_story} report the scores obtained using GPT-4o in the LLMs-as-judge setting for the evaluation of native and English stories, respectively.

\begin{table}[t]
\centering
\small
\setlength{\tabcolsep}{3pt}
\begin{tabular}{@{}lccccccccc@{}}
\toprule
& \multicolumn{3}{c}{\textbf{China}} & \multicolumn{3}{c}{\textbf{Indonesia}} & \multicolumn{3}{c}{\textbf{Egypt}} \\
\cmidrule(lr){2-4} \cmidrule(lr){5-7} \cmidrule(lr){8-10}
& CR & Fl & CO & CR & Fl & CO & CR & Fl & CO \\
\midrule
%\multicolumn{10}{c}{\textit{Llama3.1-8B-IT}} \\
Llama3.1-8B-IT& 6.0 & \textbf{7.6} & \textbf{4.8} & 6.2 & 7.1 & 4.5 & 5.9 & \textbf{7.3 }& 4.6 \\
+Mango & 6.3 & 6.6 & 4.5 & 6.7 & \textbf{6.9} & \textbf{4.8} & \textbf{6.6} & 7.2 & \textbf{4.8} \\
+CCKG & \textbf{7.9} & \textbf{6.1} & 4.2 & \textbf{6.8} & 6.7 & 4.5 & \textbf{6.6} & 6.9 & \textbf{4.8} \\
\midrule
%\multicolumn{10}{c}{\textit{Qwen2.5-7B-IT}} \\
Qwen2.5-7B-IT & 6.7 & \textbf{8.2} & \textbf{7.3} & 6.3 & \textbf{7.6} & 5.8 & 5.8 & \textbf{7.8} & 5.6 \\
+Mango & 6.8 & 8.1 & 4.9 & 6.8 & 7.2 & 4.8 & 6.6 & 7.5 & 5.2 \\
+CCKG& \textbf{7.4} & 8.0 & 6.4 & \textbf{6.9} & 7.5 & \textbf{5.9} & \textbf{6.7} & 7.4 & \textbf{5.8} \\
\midrule
%\multicolumn{10}{c}{\textit{Gemma2-9B-IT}} \\
Gemma2-9B-IT & \textbf{7.7} & 7.7 & \textbf{6.1} & 7.5 & \textbf{8.1} & \textbf{5.8} & 6.5 & \textbf{8.3} & \textbf{5.5} \\
+Mango & 7.6 & \textbf{8.0} & 5.4 & \textbf{7.7} & 7.8 & 5.2 & \textbf{7.0} & 7.7 & 4.8 \\
+CCKG & 7.6 & 7.2 & 5.2 & 7.5 & \textbf{8.1} & \textbf{5.8} & 6.9 & 7.7 & 5.2 \\
\bottomrule
\end{tabular}
\caption{GPT-4o as Judge scores across models and countries for English story generation, comparing the base setting (no augmentation) with two augmented settings: Mango and CCKG. Best scores are bolded.}
\label{tab:results_english_llm_story}
\end{table}

\begin{table}[t]
\centering
\small
\setlength{\tabcolsep}{3pt}
\begin{tabular}{@{}lccccccccc@{}}
\toprule
& \multicolumn{3}{c}{\textbf{China}} & \multicolumn{3}{c}{\textbf{Indonesia}} & \multicolumn{3}{c}{\textbf{Egypt}} \\
\cmidrule(lr){2-4} \cmidrule(lr){5-7} \cmidrule(lr){8-10}
& CR & Fl & CO & CR & Fl & CO & CR & Fl & CO \\
\midrule
%\multicolumn{10}{c}{\textit{Llama3.1-8B-IT}} \\
Llama3.1-8B-IT & 5.6 & \textbf{7.0} & 4.8 & 5.4 & 5.5 & 4.4 & 1.6 & \textbf{1.8 }& \textbf{1.6} \\
+Mango & \textbf{7.2} & 6.9 & \textbf{5.6} & 6.3 & \textbf{5.8} & \textbf{4.7} & \textbf{1.8} & \textbf{1.8} & 1.5 \\
+CCKG & \textbf{7.2} & 6.0 & 5.2 & \textbf{6.4} & 5.6 & 4.6 & \textbf{1.8} & 1.6 & 1.5 \\
\midrule
%\multicolumn{10}{c}{\textit{Qwen2.5-7B-IT}} \\
Qwen2.5-7B-IT & 6.6 & 7.2 & 6.8 & 6.2 & 6.9 & 6.4 & 3.9 & \textbf{3.4} & 3.2 \\
+Mango & 7.5 & \textbf{7.9} & 7.2 & \textbf{7.2} & 7.4 & 7.1 & 4.3 & 2.9 & \textbf{3.0} \\
+CCKG& \textbf{7.6} & 7.8 & \textbf{7.7} & \textbf{7.2} & \textbf{7.7} & \textbf{7.5} & \textbf{4.5} & 2.7 & \textbf{3.0} \\
\midrule
%\multicolumn{10}{c}{\textit{Gemma2-9B-IT}} \\
Gemma2-9B-IT & \textbf{8.6} & \textbf{9.6} & \textbf{8.8} & 8.2 & 9.3 & 8.4 & 5.8 & \textbf{5.2} & 5.1 \\
+Mango & 8.4 & 8.6 & 8.0 & \textbf{8.5} & \textbf{8.9} & 8.1 & \textbf{6.4} & 5.0 & 5.1 \\
+CCKG & 8.0 & 8.5 & 8.3 & 8.3 & \textbf{8.9} & \textbf{8.5} & 6.3 & 5.0 & \textbf{5.3} \\
\bottomrule
\end{tabular}
\caption{GPT-4o as Judge scores across models and countries for native story generation, comparing the base setting (no augmentation) with two augmented settings: Mango and CCKG. Best scores are bolded.}
\label{tab:results_native_llm_story}
\end{table}

\begin{table}[t]
\centering
\small
\begin{tabular}{@{}lccc@{}}
\toprule
\textbf{Story} & \textbf{Cultural} & \textbf{Fluency} & \textbf{Coherence} \\
\textbf{Generation} & \textbf{Relevance} & & \\
\midrule
\multicolumn{4}{@{}l}{\cellcolor{blue!7}\textit{Indonesia}} \\
\addlinespace[3pt]
Indonesian & 0.7 & 0.8 & 0.7 \\
English    & 0.4 & 0.1 & 0.0 \\
\midrule
\multicolumn{4}{@{}l}{\cellcolor{blue!7}\textit{China}} \\
\addlinespace[3pt]
Chinese & 0.7 & 0.8 & 0.7 \\
English & 0.5 & 0.1 & 0.1 \\
\midrule
\multicolumn{4}{@{}l}{\cellcolor{blue!7}\textit{Egypt}} \\
\addlinespace[3pt]
MSA     & 0.9 & 0.9 & 0.9 \\
English & 0.4 & $-0.6$ & $-0.1$ \\
\bottomrule
\end{tabular}
\caption{Pearson correlation between LLM evaluations and human annotators (Annotator~1 and Annotator~2) for the story-generation task across English and native-language settings.}
\label{tab:llm-human-agreement}
\end{table}

\subsection{Assessment of Cultural Generalization and Stereotyping}
\label{sec:stereo}

We conducted a qualitative assessment by two native speakers of 30 assertions related to the culturally sensitive topics of \textit{wedding} and \textit{death} across four countries, namely Egypt, China, Indonesia, and Japan, in both English and native-language.
For each setting, we applied the following evaluation instruction:
\begin{quote}
\textit{Assess whether the text contains overly generalized, stereotypical, or unfair assumptions about individuals within the target culture, including but not limited to those based on gender roles, religion, ethnicity, regional affiliation, or socioeconomic class.}
\end{quote}
Table \ref{tab:cultural-generalization} reports the average percentage of annotator judgments (``Yes'', ``No'', and ``Unsure''). Across most countries and languages, the majority of extracted assertions were judged as not containing overly generalized or stereotypical assumptions. In our extractions, we also observed that the model frequently uses modal language such as `might` and `will likely` to describe plausible reactions and responses of people in the given situations.
Nonetheless, stereotypes and bias in LLMs are serious concerns, so we still recommend that future work include more detailed and large-scale audits on more topics.

\begin{table}[t]
\centering
\small
\setlength{\tabcolsep}{2pt}
\begin{tabular}{lccc}
\toprule
 & \textbf{No (\%)} & \textbf{Yes (\%)} & \textbf{Unsure (\%)} \\
\midrule
Egypt (EN)        & 100.0 & 0 & 0.0 \\
Egypt (MSA)       & 93.3 &  3.3 &  3.3 \\
\midrule
Indonesia (IND)   & 90.0 &  5.0 &  5.0 \\
Indonesia (EN)    & 98.3 &  1.7 &  0.0 \\
\midrule
Japan (EN)        & 91.7 &  8.3 &  0.0 \\
Japan (JAP)       & 80.0 & 20.0 &  0.0 \\
\midrule
China (EN)        &100.0 &  0.0 &  0.0 \\
China (CHI)       &100.0 &  0.0 &  0.0 \\
\bottomrule
\end{tabular}
\caption{
Average percentage of human judgments assessing whether extracted assertions related to \textit{wedding} and \textit{death} contain overly generalized or stereotypical cultural assumptions.
}
\label{tab:cultural-generalization}
\end{table}

\subsection{MCQA Results with Native Prompts}
\label{sec:Mcqa_native}
Table \ref{tab:results_native_mcq} reports accuracy scores on the ArabCulture and IndoCulture benchmarks for the MCQA task using native prompts. The results follow the same trends observed with English prompts, as discussed in the main text.

\begin{table*}[t]
\centering
\small
\resizebox{\textwidth}{!}{
\setlength{\tabcolsep}{0.7mm}
\begin{tabular}{lrrrrrrrcrrrrrrr}
\toprule
\multicolumn{8}{c}{\cellcolor{blue!7} \textbf{IndoCulture}} & \multicolumn{7}{c}{\cellcolor{red!7}\textbf{ArabCulture}} \\
\cmidrule(lr){2-8}\cmidrule(lr){9-15}
\multirow{2}{*}{\textbf{Models}} 
  & \multicolumn{2}{c}{\cellcolor{blue!7}\textbf{Before Aug}} 
  & \multicolumn{5}{c}{\cellcolor{blue!7}\textbf{After Aug}} 
  & \multicolumn{2}{c}{\cellcolor{red!7}\textbf{Before Aug}} 
  & \multicolumn{5}{c}{\cellcolor{red!7}\textbf{After Aug}} \\
\cmidrule(lr){2-3} \cmidrule(lr){4-8}
\cmidrule(lr){9-10} \cmidrule(lr){11-15}
& \cellcolor{blue!7}\textbf{Base} & \cellcolor{blue!7}\textbf{CoT} 
& \cellcolor{blue!7}\textbf{Mango} 
& \cellcolor{blue!7}\textbf{E-Asrt} & \cellcolor{blue!7}\textbf{E-Path} & \cellcolor{blue!7}\textbf{N-Asrt} & \cellcolor{blue!7}\textbf{N-Path}
& \cellcolor{red!7}\textbf{Base} & \cellcolor{red!7}\textbf{CoT} 
& \cellcolor{red!7}\textbf{Mango} 
& \cellcolor{red!7}\textbf{E-Asrt} & \cellcolor{red!7}\textbf{E-Path} & \cellcolor{red!7}\textbf{N-Asrt} & \cellcolor{red!7}\textbf{N-Path} \\
\midrule
Qwen2.5-0.5B     & 40.8 & 37.3 & 40.9 & \textbf{41.6} & 38.5 & 40.6 & 39.0 & 34.3 & 34.3 & 34.3 & 34.4 & 34.3 & 34.2 & \textbf{34.5} \\
Qwen2.5-1.5B     & 38.8 & 33.7 & 44.7 & \textbf{44.8} & 42.1 & 40.5 & 38.6 & \textbf{38.8 }& 35.2 & 35.5 & 37.5 & 36.2 & 37.2 & 36.6 \\
Qwen2.5-3B       & 52.5 & 53.1 & \textbf{54.8} & 53.9 & 52.7 & 51.9 & 51.0 & 37.5 & 36.4 & 37.9 & 36.9 & 36.7 & \textbf{39.3} & 37.8 \\
Qwen2.5-7B       & 58.5 & 56.6 & 59.4 & 59.2 & \textbf{59.8} & 59.3 & 59.5 & \textbf{47.9} & 35.4 & 38.7 & 40.5 & 39.2 & 37.5 & 38.7 \\
Gemma2-2B         & 34.5 & 35.9 & 38.1 & \textbf{40.0} & 38.6 & 39.1 & 36.6 & 34.3 & 34.3 & 34.3 & 34.3 & 34.3 & 34.3 & 34.3 \\
Gemma2-9B         & 65.4 & 42.2 & 66.6 & 65.6 & \textbf{67.5} & 65.7 & 67.4 & 35.6 & 36.6 & 35.5 & 37.3 & 35.3 & \textbf{38.1} & 34.8 \\
Llama3.2-1B       & 56.7 & 56.0 & 54.8 & 55.6 & 55.0 & 55.7 & \textbf{57.1} & 33.9 & \textbf{34.2} & 33.8 & 34.1 & 34.1 & 33.8 & 34.1 \\
Llama3.1-8B       & 32.7 & \textbf{35.6} & 32.7 & 32.7 & 32.7 & 32.7 & 32.7 & 34.3 & 34.3 & 34.2 & 34.3 & 34.2 & 34.0 & 34.2 \\
Llama3.2-3B       & 47.1 & 44.1 & 46.9 & 47.4 & 46.7 & \textbf{48.3} & 47.3 & 34.3 & 34.1 & \textbf{34.4} & \textbf{34.4} & 34.3 & 34.3 & 34.3 \\
\midrule
Gemma2-9B-IT      & 57.4 & 56.7 & 57.1 & 56.2 & 57.8 & \textbf{58.5} & 58.1 & 34.3 & \textbf{34.8} & 34.4 & 34.3 & 34.3 & 34.3 & 34.3 \\
Qwen2.5-7B-IT     & 66.1 & \textbf{67.2} & 64.6 & 65.5 & 65.6 & 65.4 & 65.7 & \textbf{39.5} & 34.3 & 35.6 & 36.8 & 38.0 & 35.8 & 36.2 \\
Llama3.1-8B-IT     & 53.4 & 48.9 & 55.2 & 55.6 & 53.6 & 55.7 & \textbf{55.9} & 37.9 & 34.2 & 34.3 & 34.3 & \textbf{34.4} & 34.3 & 34.4 \\
\midrule
{Avg $\Delta$} & NA & $-$3 & 1 & \textbf{1.2 }& 0.6 & 0.8 & 0.4 &NA & $-$2.1 & $-$1.7 & $-$1.2 & $-$1.5 & $-$1.3 & $-$1.5\\
%\textbf{Avg} & 50.3 & 47.3 & 51.3 & \textbf{51.5} & 50.9 & 51.1 & 50.7
%                 & \textbf{36.9} & 34.8 & 35.2 & 35.7 & 35.4 & 35.6 & 35.4 \\
\bottomrule
\end{tabular}
}
\caption{Accuracy comparison on MCQA across different methods with native prompts. \textbf{E-Asrt}: CCKG English Assertions, \textbf{E-Path}: CCKG English Paths, \textbf{N-Asrt}: CCKG Native-language (in Arabic or Indonesian) Assertions, \textbf{N-Path}: CCKG Native-language (in Arabic or Indonesian) Paths.  ``Avg $\Delta$'' denotes the average improvement over the baseline. Best results per model are bolded. }
\label{tab:results_native_mcq}
\end{table*}

\subsection{Sentence Completion Results with Native Prompts}
\label{sec:sentence_completion_native}

Table \ref{tab:similarity_bert_results_completion_native} presents BERT F1 and sentence similarity scores on the ArabCulture and IndoCulture benchmarks for the sentence completion task with native prompts, again showing the same trends as with English prompts.

\begin{table}[ht!]
\centering
\resizebox{0.99\columnwidth}{!}{%
\setlength{\tabcolsep}{1.8pt}
\begin{tabular}{lccccccc}
\toprule
\multirow{3}{*}{\textbf{Models}} & \multirow{3}{*}{\begin{tabular}[c]{@{}c@{}}\textbf{Before}\\\textbf{Aug}\end{tabular}} & \multirow{3}{*}{\textbf{+Mango}} & \multicolumn{4}{c}{\textbf{+CCKG Methods}} \\
\cmidrule(lr){4-7}
& & & \textbf{E-Asrt} & \textbf{E-Path} & \textbf{N-Asrt} & \textbf{N-Path} \\
\midrule
\multicolumn{7}{c}{\cellcolor{red!7}\textbf{ArabCulture - Avg Sentence Similarity}} \\
\midrule
Gemma2-9B-IT & 31.6 & 32.1 & 35.0 & 32.7 & \textbf{35.4} & 33.2 \\
Qwen2.5-7B-IT & 40.2 & 42.3 & 43.9 & 41.3 & \textbf{44.7} & 41.9 \\
Llama3.1-8B-IT & 27.3 & \textbf{32.4} & 30.3 & 28.0 & 28.5 & 27.2 \\
\textbf{Avg} & 33.0 & 35.6 & 36.4 & 34.0 & \textbf{36.2} & 34.1 \\
\midrule
\multicolumn{7}{c}{\cellcolor{blue!7}\textbf{IndoCulture - Avg Sentence Similarity}} \\
\midrule
Gemma2-9B-IT & 31.3 & 33.2 & 34.0 & 34.0 & 34.6 & \textbf{35.0} \\
Qwen2.5-7B-IT & 39.8 & 41.1 & \textbf{42.0} & 41.7 & 41.5 & 34.4 \\
Llama3.1-8B-IT & 31.1 & 33.6 & 33.8 & 34.4 & 35.0 & \textbf{41.6} \\
\textbf{Avg} & 34.1 & 35.9 & 36.6 & 36.7 & \textbf{37.1} & 37.0 \\
\midrule
\multicolumn{7}{c}{\cellcolor{red!7}\textbf{ArabCulture - Avg BERT Score F1}} \\
\midrule
Gemma2-9B-IT & 68.1 & 68.2 & 68.4 & 68.4 & \textbf{68.6} & 68.3 \\
Qwen2.5-7B-IT & 68.3 & 68.7 & 68.8 & 67.7 & \textbf{69.5} & 68.7 \\
Llama3.1-8B-IT & 65.0 & \textbf{66.0} & 65.8 & 65.0 & 65.6 & 65.3 \\
\textbf{Avg} & 67.1 & 67.6 & 67.7 & 67.0 & \textbf{67.9} & 67.4 \\
\midrule
\multicolumn{7}{c}{\cellcolor{blue!7}\textbf{IndoCulture - Avg BERT Score F1}} \\
\midrule
Gemma2-9B-IT & 70.5 & 70.8 & 71.1 & 71.1 & 71.3 & \textbf{71.4} \\
Qwen2.5-7B-IT & 71.9 & 71.9 & 71.9 & 71.9 & \textbf{72.0} & 70.1 \\
Llama3.1-8B-IT & 69.6 & 70.0 & 70.2 & 70.0 & 70.2 & \textbf{72.1} \\
\textbf{Avg} & 70.7 & 70.9 & 71.1 & 71.0 & \textbf{71.2} & \textbf{71.2} \\
\bottomrule
\end{tabular}
}
\caption{Performances for sentence similarity and BERT score F1 with native prompts. \textbf{E-Asrt}: CCKG English Assertions, \textbf{E-Path}: CCKG English Paths, \textbf{N-Asrt}: CCKG Native-language (in Arabic or Indonesian) Assertions, \textbf{N-Path}: CCKG Native-language (in Arabic or Indonesian) Paths. Best results per row are bolded.}
\label{tab:similarity_bert_results_completion_native}
\end{table}

\subsection{Results by Criterion: English vs. Native Story Generation}
\label{sec:english_native_wise}
\begin{figure*}[t]
  \centering
  \includegraphics[width=0.95\textwidth]{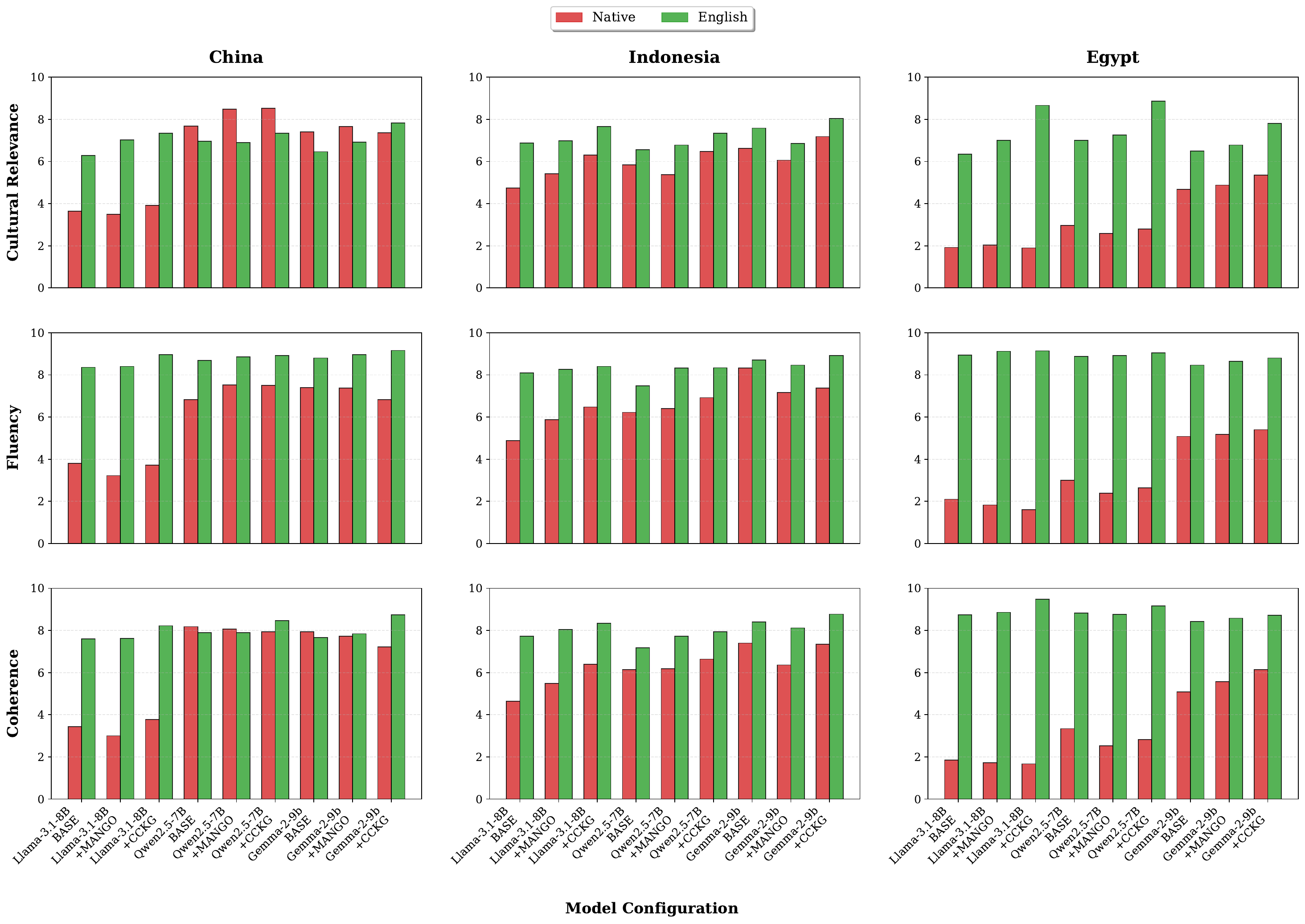}
  \caption{Average scores per evaluation metric for each country and model.
 }
  \label{fig:country_model_wise_results}
\end{figure*}

Figure \ref{fig:country_model_wise_results} illustrates the average score for each evaluation metric, broken down by country and model.

\section{Prompts}
This section presents all the prompts used in our experiments. All the translations were produced by native speakers of the country.

\subsection{Prompts for CCKG}
\label{sec:promtp_cckg}

\begin{figure*}[t]
  \centering
  \includegraphics[width=\textwidth]{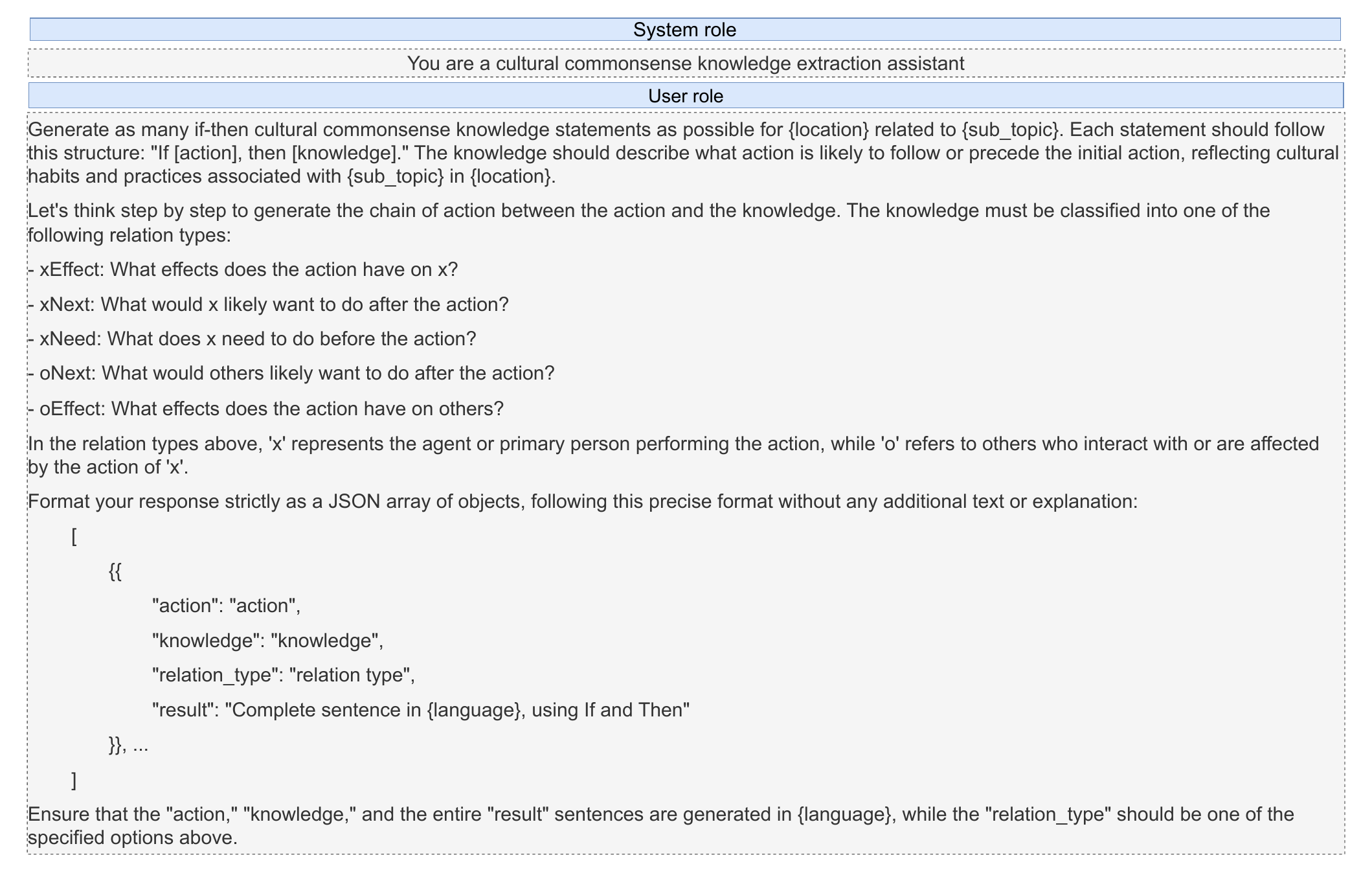}
  \caption{Prompt used in the initial generation step of CCKG. The variables \texttt{sub\_topic}, \texttt{location}, and \texttt{language} are replaced with their corresponding values (sub-topic, country, and language), expressed in English when the KB is generated in English and in the native language otherwise.
 }
  \label{fig:initial-prompt}
\end{figure*}

\begin{figure*}[t]
  \centering
  \includegraphics[width=\textwidth]{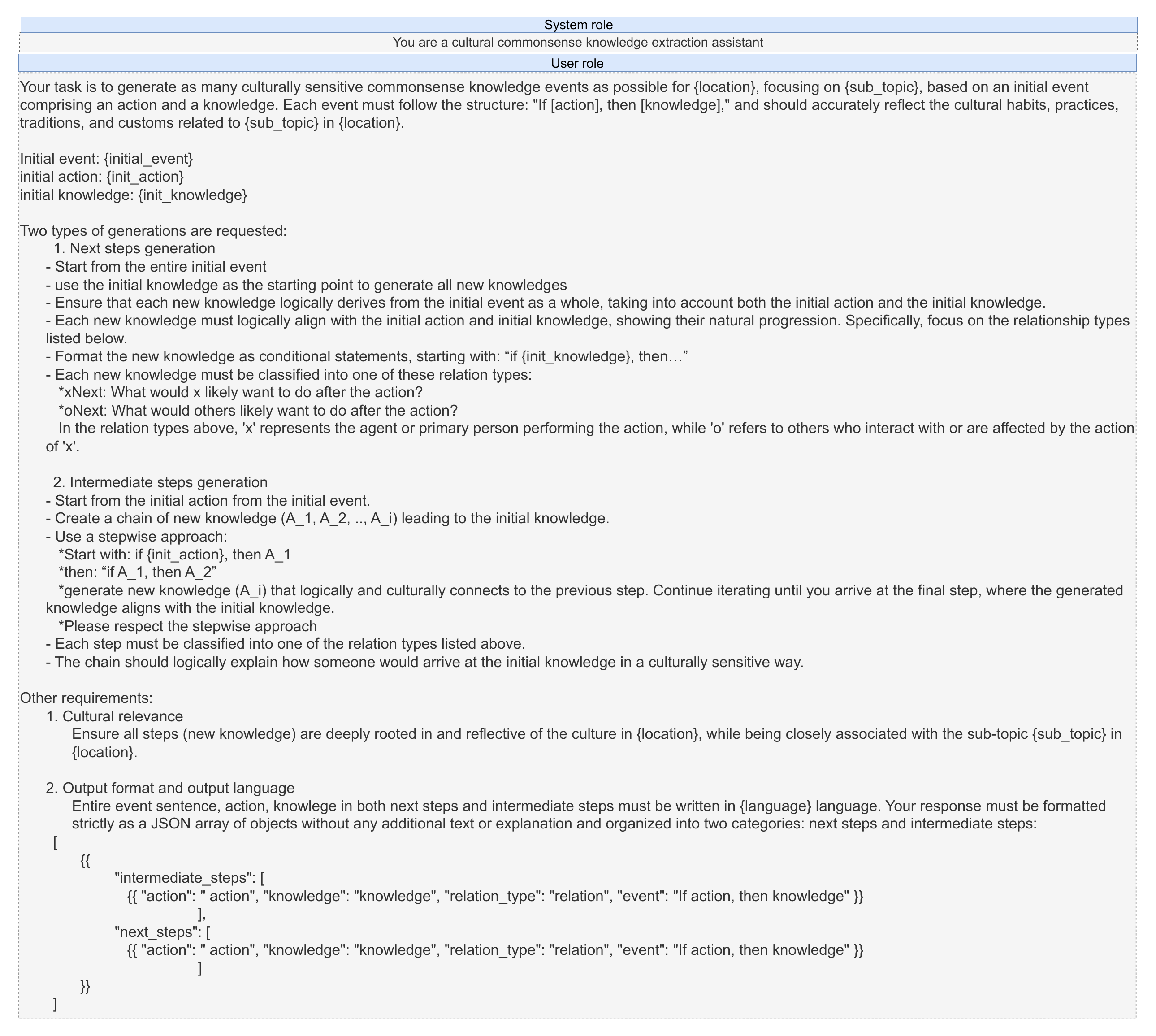}
  \caption{Prompt used in the iterative expansion step of CCKG. The variables \texttt{sub\_topic}, \texttt{location}, and \texttt{language} are replaced with the corresponding values, expressed in English when the KB is generated in English and in the native language otherwise. The variables \texttt{initial\_event}, \texttt{init\_action}, and \texttt{init\_knowledge} are instantiated from the initial assertion: for an assertion ``if \textit{action\_1}, then \textit{action\_2},'' \texttt{initial\_event} is the full assertion, \texttt{init\_action} is \textit{action\_1}, and \texttt{init\_knowledge} is \textit{action\_2}.
}
  \label{fig:expansion-prompt}
\end{figure*}

Figure~\ref{fig:initial-prompt} illustrates the prompt employed during the initial generation phase of our algorithm, while Figure~\ref{fig:expansion-prompt} shows the prompt used in the iterative expansion phase. Both prompts are applied verbatim when constructing CCKG in English. For cases where CCKG is generated in a non-English language, the same prompts are translated into the target language. In our experiments, this includes Modern Standard Arabic (MSA), Chinese, Japanese, and Bahasa Indonesian.

\begin{figure*}[t]
  \centering
  \includegraphics[width=\textwidth]{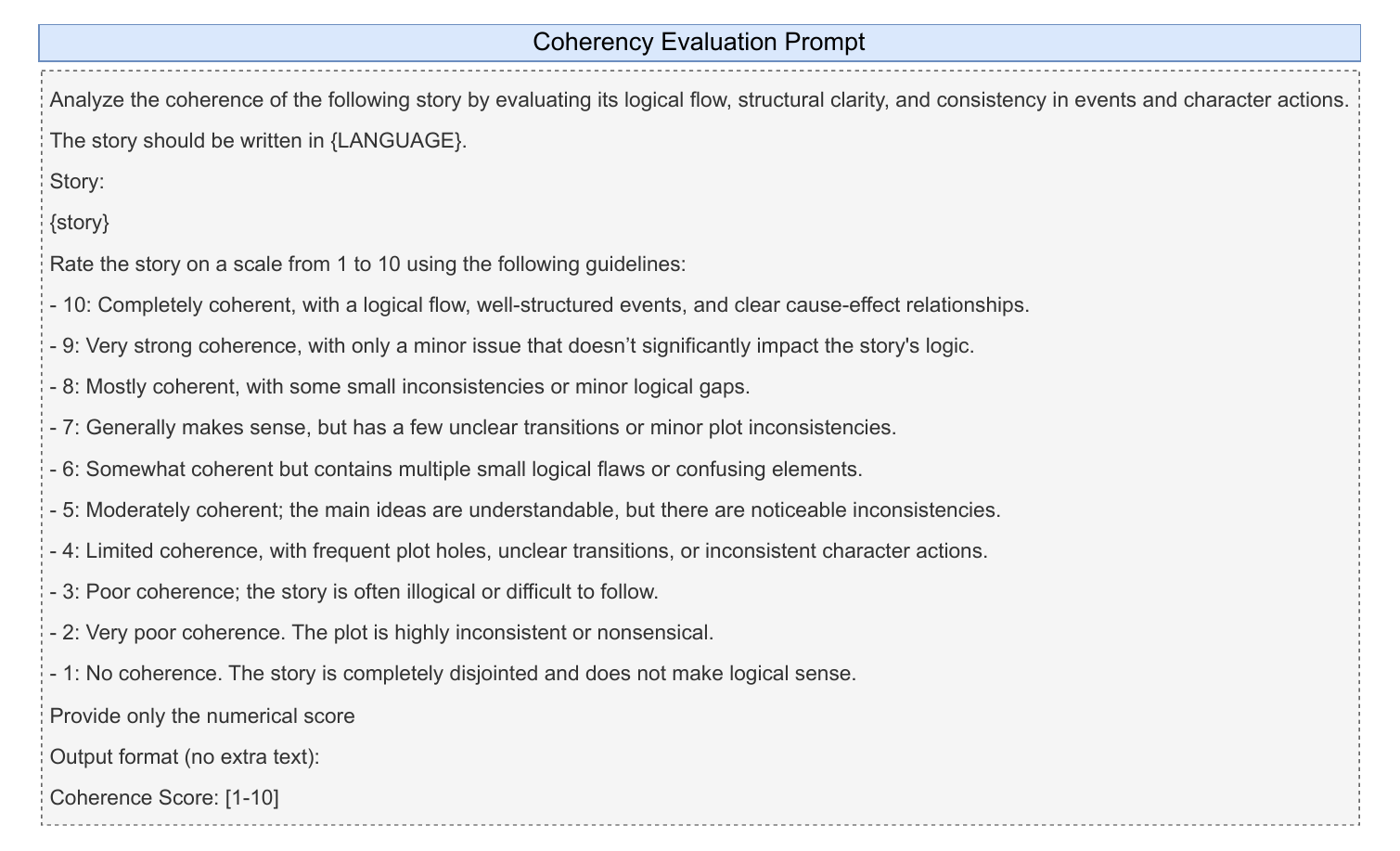}
  \caption{Prompt used to evaluate coherence in the story generation task.}
  \label{fig:eval-coherence}
\end{figure*}

\begin{figure*}[t]
  \centering
  \includegraphics[width=\textwidth]{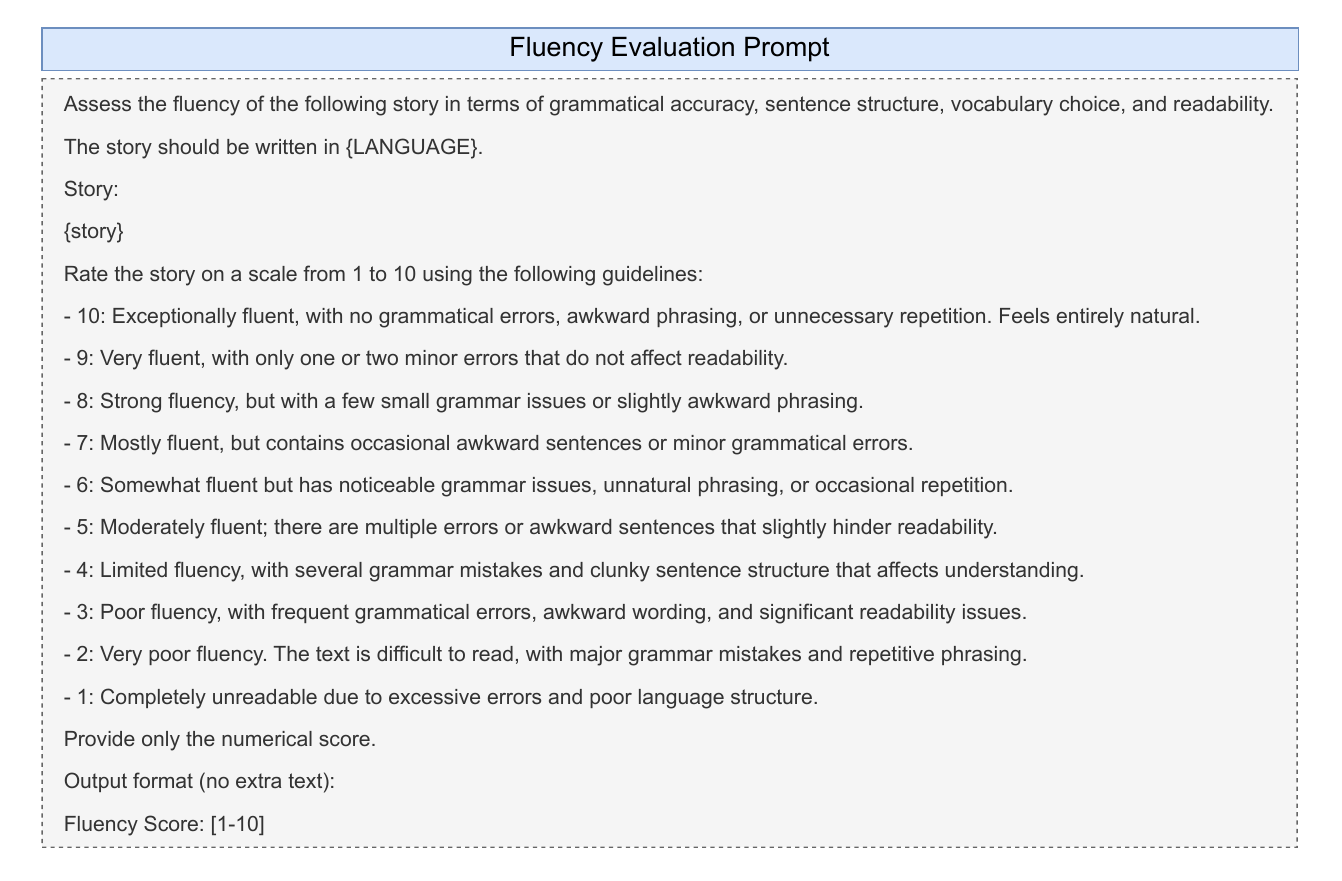}
  \caption{Prompt used to evaluate fluency in the story generation task.}
  \label{fig:eval-fluency}
\end{figure*}

\begin{figure*}[t]
  \centering
  \includegraphics[width=\textwidth]{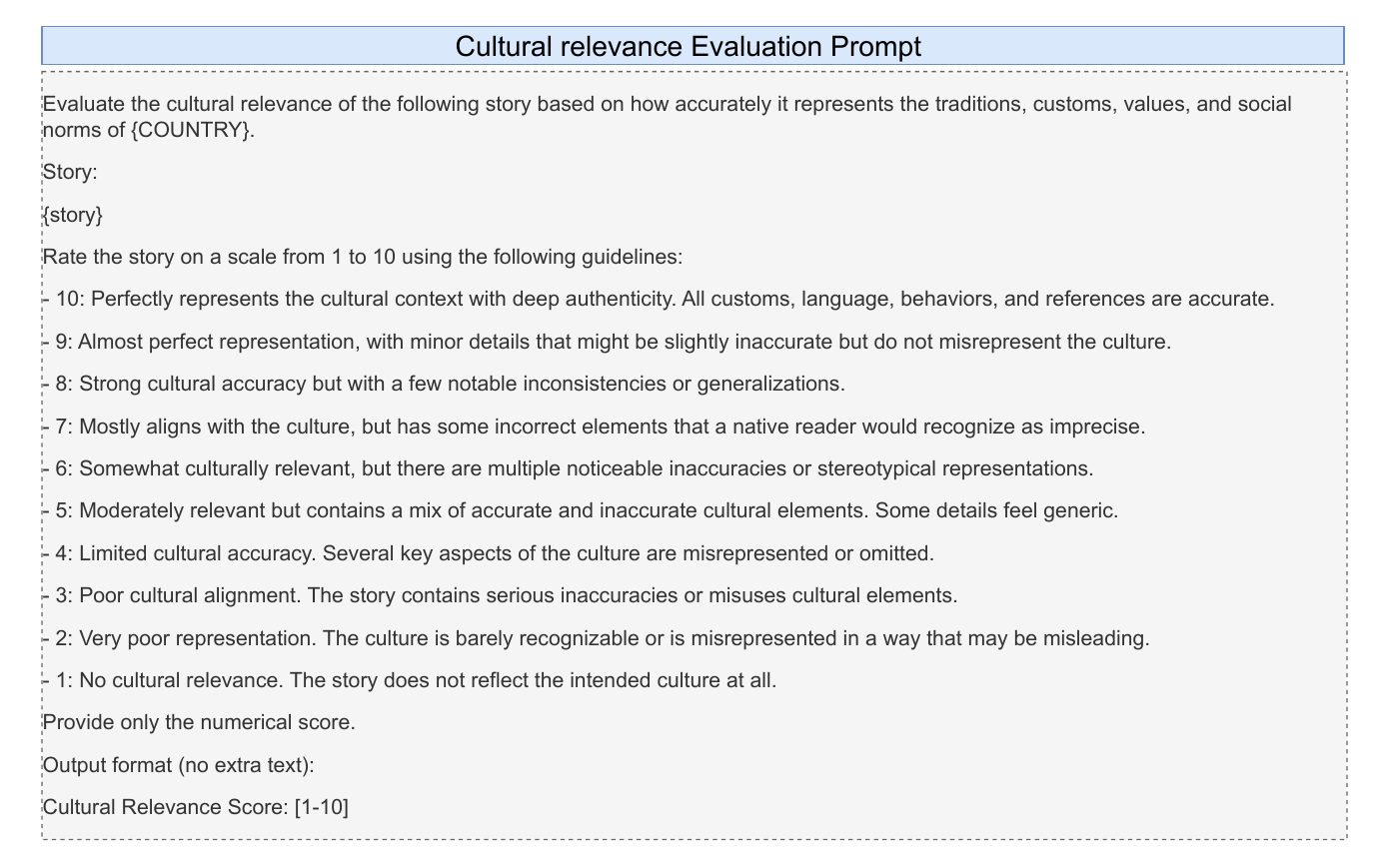}
  \caption{Prompt used to evaluate cultural relevancy in the story generation task.}
  \label{fig:eval-cultural}
\end{figure*}

\begin{figure*}[t]
  \centering
  \includegraphics[width=0.8\textwidth]{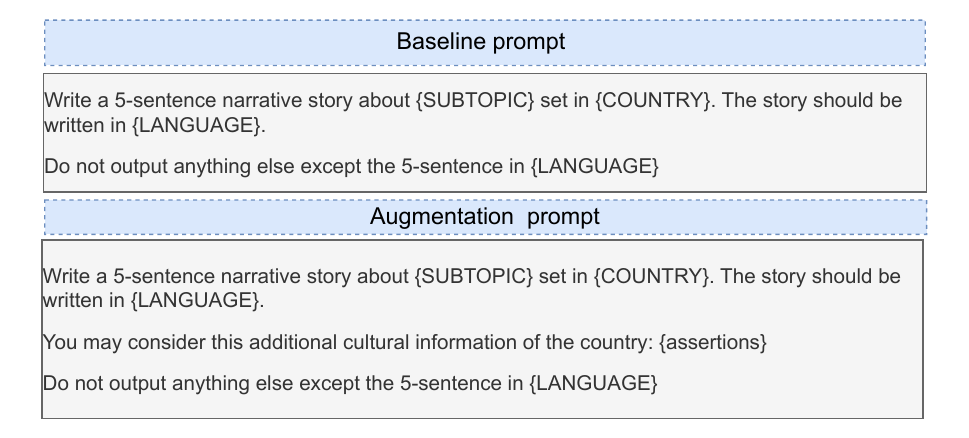}
  \caption{Story generation prompts.}
  \label{fig:story-prompts}
\end{figure*}

\subsection{Prompts for Story Generation}
\label{sec:story_gen_prompt}
Figure~\ref{fig:story-prompts} presents the base and augmentation prompts used for story generation in both English and the native languages. The variables \texttt{SUBTOPIC}, \texttt{COUNTRY}, \texttt{LANGUAGE}, and the assertions are replaced with their corresponding values. When the story is generated in English, the variable \texttt{LANGUAGE} is set to ``English''; otherwise, it is set to the respective native language.  

\subsection{Prompts for Evaluation with LLMs-as-Judge}
\label{sec:Llm_judges_prompt}

Figures~\ref{fig:eval-cultural}, \ref{fig:eval-fluency}, and \ref{fig:eval-coherence} show the prompts used to evaluate cultural relevance, fluency, and coherence, respectively. In all cases, the \texttt{story} and \texttt{country} variables are replaced with their corresponding values. The same prompts were also provided to human annotators for the human evaluation of stories generated in both English and native languages. 

\subsection{Prompts for MCQA and SENTENCE COMPLETION}
For MCQA and sentence completion tasks, we use the original benchmark prompts.  
\end{document}